\documentclass[10pt,twocolumn,letterpaper]{article}

\usepackage[pagenumbers]{cvpr} 

\usepackage{graphicx}
\usepackage{amsmath}
\usepackage{amssymb}
\usepackage{booktabs}
\usepackage{multirow}
\usepackage{multicol}
\usepackage{dsfont}
\usepackage{colortbl}
\usepackage{lipsum}
\usepackage{xspace}
\usepackage{makecell}
\usepackage{enumitem}
\usepackage[table,xcdraw]{xcolor}
\usepackage[section]{placeins}
\usepackage{float}

\usepackage{subfloat}

\usepackage{algorithm}
\usepackage{algorithmicx}
\usepackage{algorithm} 
\usepackage[noend]{algpseudocode} 
\usepackage[none]{hyphenat}

\definecolor{mygray}{gray}{.9}
\definecolor{mypink}{rgb}{.99,.91,.95}
\definecolor{mycyan}{cmyk}{.3,0,0,0}

\setitemize{noitemsep,topsep=0pt,parsep=0pt,partopsep=0pt}

\newcommand{\vpara}[1]{\vspace{0.07in}\noindent\textbf{#1}\xspace}

\newcommand{\gps}{\text{AutoGPart }}   
\newcommand{\gpsss}{\text{AutoGPart}} 
\newcommand{\eat}[1]{}

\usepackage[pagebackref,breaklinks,colorlinks]{hyperref}

\usepackage[capitalize]{cleveref}
\crefname{section}{Sec.}{Secs.}
\Crefname{section}{Section}{Sections}
\Crefname{table}{Table}{Tables}
\crefname{table}{Tab.}{Tabs.}

\def\cvprPaperID{2539} 
\def\confName{CVPR}
\def\confYear{2022}

\begin{document}

\title{
AutoGPart: 
Intermediate Supervision Search \\ for Generalizable 3D Part Segmentation 
}

\author{
\vspace{-6pt}
Xueyi Liu$^1$ 
\and
Xiaomeng Xu$^1$ 
\and
Anyi Rao$^2$ 
\and
Chuang Gan$^3$ 
\and
Li Yi$^{1,4}$ 
\and
$^1$ Tsinghua University \hspace{0.5cm} $^2$ The Chinese University of Hong Kong \hspace{0.5cm} $^3$ MIT-IBM Watson AI Lab \\ $^4$ Shanghai Qi Zhi Institute
}
\maketitle

 
\begin{abstract}
Training a generalizable 3D part segmentation network is quite challenging but of great importance in real-world applications. 
To tackle this problem, some works design task-specific solutions by translating human understanding of the task to machine's learning process, which faces the risk of missing the optimal strategy since machines do not necessarily understand in the exact human way. 
Others try to use conventional task-agnostic approaches designed for domain generalization problems with no task prior knowledge considered. 
To solve the above issues, we propose AutoGPart, a generic method enabling training generalizable 3D part segmentation networks with the task prior considered. 
AutoGPart builds a supervision space with geometric prior knowledge encoded,
and lets the machine to search for the optimal supervisions from the space for a specific segmentation task automatically.
Extensive experiments on three generalizable 3D part segmentation tasks are conducted to demonstrate the effectiveness and versatility of AutoGPart. 
We demonstrate that the performance of segmentation networks using simple backbones can be significantly improved when trained with supervisions searched by our method. Project page: \href{https://autogpart.github.io}{https://autogpart.github.io}.
\end{abstract}
\section{Introduction}
\label{sec:intro}

\begin{figure}[ht]
\vspace{-5pt}
  \centering
   \includegraphics[width=0.90\linewidth]{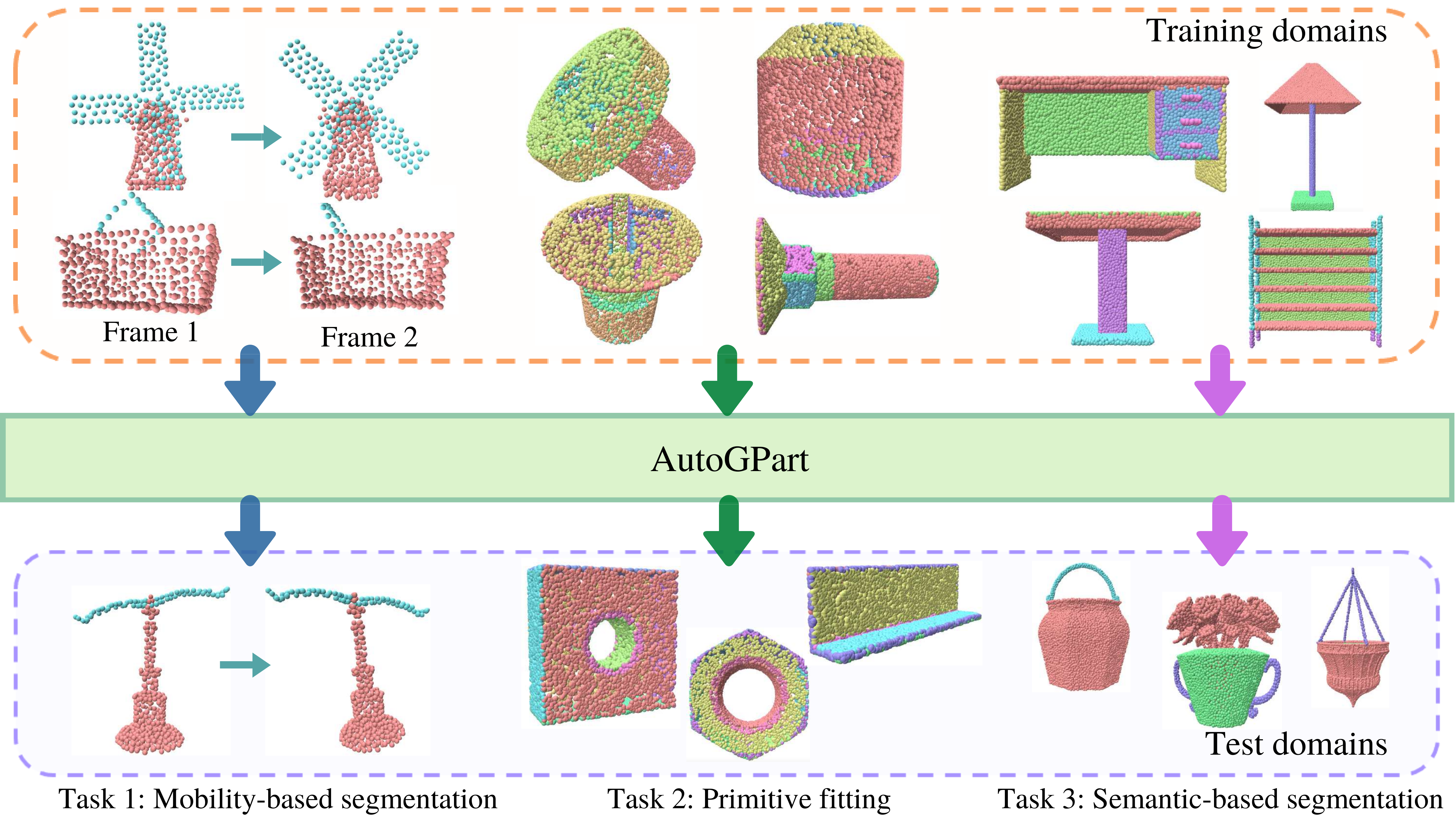}
  \vspace{-6pt}
  \caption{
  \footnotesize
\emph{AutoGPart} automatically 
finds 
real 
part cues for different segmentation tasks and supports 
 simple point cloud processing backbones (\eg  PointNet++~\cite{qi2017pointnetpp})  to successfully parse shapes from novel categories.
}
\vspace{-12pt}
  \label{fig_task_overview}
\end{figure}


Humans parse objects into parts for a deeper understanding of semantics, functionality, mobility as well as the fabrication process. It is natural to consider equipping machines with such part-level understanding. Over the past few years, there has been an increasing interest in 3D part segmentation tasks to support various applications~\cite{mo2019partnet,li2019supervised,huang2021primitivenet,Yan_2021_ICCV}
in vision, graphics, and robotics. Thanks to the availability of large-scale 3D part datasets~\cite{yi2016scalable,mo2019partnet} and the development of 3D deep learning techniques, these methods achieve impressive part segmentation results when a test shape comes from the same distribution as the training set. However, they usually suffer from a huge performance drop when it comes to parsing 3D shapes from a novel distribution, \eg from a different semantic category.

In this work, we focus on generalizable 3D part segmentation tasks. The goal is to learn the essence of parts on some training sets and to be able to generalize well to shapes from novel distributions. A main challenge comes from the versatile cues defining parts. Shape parts can be defined via many cues like geometric primitive fitting~\cite{Yan_2021_ICCV,li2019supervised,huang2021primitivenet}, rigid motions~\cite{yi2018deep,hu2012co,hu2017learning}, and semantic prior~\cite{luo2020learning,mo2019partnet,liu2017sgn}. The same shape can be segmented into different parts based upon different cues for the seek of various applications. Therefore for a specific application, other than understanding input shapes, a network also needs to figure out the exact cues defining parts from the training data. This is where it becomes tricky. Some of these cues are more salient than others, making them easier to be captured. When these cues heavily correlate with the real part cue in the training domain, they could easily become shortcut features~\cite{geirhos2020shortcut,geirhos2018imagenet,ilyas2019adversarial} biasing the network and hindering the generalizability, as observed in~\cite{luo2020learning}.

To cope with the above challenges, existing works usually focus on a specific type of parts and incorporate part-type priors towards generalizable 3D part segmentation. For example, a modularized design has been adopted to explicitly extract motion flow and the rigid motion of local patches so that generalizable motion-based part segmentation can be achieved in~\cite{yi2018deep}. Similarly, the networks are guided explicitly to extract primitive parameters in order to segment primitive parts in~\cite{li2019supervised,Yan_2021_ICCV}.
Finding such explicit part type priors requires a deep understanding of the target task, which might not always be available. Even after a huge amount of trail-and-errors in expert designs, such approaches are still not guaranteed to fully grasp the essence of parts. This is because human understandings might not align well with the way that machines prefer to follow
~\cite{funke2020notorious,lapuschkin2019unmasking,zhang2019dissonance}.

Another line of works from the machine learning community studies generic domain generalization algorithms to help a network handle out-of-distribution samples. However, without considering any geometric prior which is important for part segmentation tasks, they fail to improve crucial parts of the model but merely focus on generic regularization strategies such as making gradients from 
multiple domains consistent with each other~\cite{Mansilla_2021_ICCV}. 

In contrast, we present \emph{AutoGPart}, a generalizable 3D part segmentation technique that could be applied to any type of 3D parts. Our key observation is that the generalizability of a 3D part segmentation network is largely hindered by shortcut features~\cite{geirhos2020shortcut}. These shortcut features tend to be quite salient and closely correlated with segmentation labels in some training domains. Through finding proper 
intermediate supervisions 
that are more closely related with real part cues 
and loosely correlated with shortcut features, 
we can downweight the influence of shortcut features by jointly training the network with segmentation and the intermediate supervisions.

We design a parametric model to depict the distribution of possible intermediate supervisions. And we inject geometric priors in the model space such that these supervisions are part-aware and geometrically-discriminative. In addition, we present an automatic way for optimizing the distribution of these supervisions 
for a supervised part segmentation network.
A beam search-like strategy is then applied to greedily generate some suitable supervisions from the distribution. With the additional supervision, task-specific part cues can be automatically discovered without experts' trail-and-errors.
Comprehensive experiments on three generalizable 3D part segmentation tasks demonstrate the effectiveness of \gpsss,
including 4.4\% absolute Segmentation Mean IoU improvement on mobility-based part segmentation , 4.2\% on  primitive fitting and 0.7\% absolute Mean Recall~\cite{luo2020learning} improvement on semantic-based part segmentation.  

To summarize, our contributions are threefold:
1) We present a generic method to improve the generalizability of 3D part segmentation networks via automatically discovered intermediate supervisions and the method is suitable for different part definitions. 
2) We design a parametric model to depict the distribution of useful intermediate supervisions with geometric prior encoded. 
3) We propose an automatic search algorithm to find the proper intermediate supervision given a specific part segmentation task.

\section{Related Work}
\noindent\textbf{Domain Generalization.}
As an important technique to handle out-of-distribution scenarios, 
existing domain generalization approaches can be categorized into four streams: 
1) learning domain invariant features from multiple source domains aiming to minimize the difference between source domains\cite{pmlr-v28-muandet13,li2018domain,li2018deep}, meta-learning\cite{li2018learning,li2019episodic}, and other model-agnostic strategies\cite{Mansilla_2021_ICCV};
2) data augmentation algorithms to simulate domain shift \cite{zhou2021domain}; 
3) ensemble learning techniques that train domain-specific models and uses their ensemble for inference\cite{xu2014exploiting,zhou2021domain};
4) automated machine learning (AutoML) techniques that aim to automatically search for data augmentation\cite{cubuk2019autoaugment,li2020dada} or neural architecture\cite{bai2021ood,pmlr-v70-cortes17a} which can achieve optimal generalization performance. 
Different from previous AutoML strategies, we propose to search for geometric and part-aware features for intermediate supervisions. 
Such features are invariant across shapes from different distributions, indicating the superiority of our method to improve the generalizability of the network. 

\vspace{2pt}
\noindent\textbf{3D Part Segmentation.}
Mobility-based part segmentation, primitive fitting, and semantic-based part segmentation are three important and representative 3D part segmentation tasks, 
on which we conduct experiments to prove the effectiveness and versatility of \gps.
1) \emph{Mobility-based part segmentation} aims to parse articulated input objects into rigidly moving parts.
A number of previous works have devoted into it based on traditional techniques such as clustering and co-segmentation~\cite{tzionas2016reconstructing,yuan2016space}, or deep 3D neural networks~\cite{yi2018deep,hu2017learning}. 
2) \emph{Primitive fitting} addresses the task of clustering input points and fitting them with geometric primitives. Standard solutions include RANSAC\cite{schnabel2007efficient}, region growing\cite{marshall2001robust}, supervised learning~\cite{li2019supervised,Yan_2021_ICCV,huang2021primitivenet} and unsupervised learning~\cite{tulsiani2017learning,fang2018planar}.
Recently, \cite{li2019supervised} proposes an end-to-end neural network that takes point clouds as input and predicts a varying number of primitives. \cite{Yan_2021_ICCV} further uses hybrid feature representations to separate points of different primitives. 
3) \emph{Semantic-based part segmentation} detects and delineates each distinct object of interest.
Conventional approaches rely on manual design on geometric constraints, including K-means~\cite{shlafman2002metamorphosis}, graph cuts~\cite{golovinskiy2008randomized} and spectral clustering~\cite{liu2004segmentation}. 
As for learning-based methods~\cite{qi2017pointnet,yi2019gspn},
although they achieve state-of-the-art performance on seen categories, 
it is hard for them to parse shapes from unseen categories 
due to category-variant information that is easy to take than real cues defining parts~\cite{luo2020learning}. 
Different from previous task-specific methods that employ sophisticated frameworks and delicately designed supervisions,
we demonstrate that simple learning frameworks equipped with the intermediate supervisions searched from \gps 
can achieve comparable or even better performance than existing task-specific designs on the three tasks. 

\vspace{2pt}
\noindent\textbf{Auxiliary Supervision.}
Adding auxiliary supervisions is a common strategy to reguralize the learning process for performance improvement.
Among them, weight decay is a widely used technique~\cite{krogh1992simple,loshchilov2017decoupled}. 
Besides, multi-task learning~\cite{crawshaw2020multi,zhang2017survey,he2017mask,ren2015faster} trains a network to solve multiple tasks by sharing parameters between different tasks, where the interested branches acquire benefits from other branches. Deeply Supervise~\cite{lee2015deeply} and Inception~\cite{szegedy2015going} explore how to add auxiliary supervisions on hidden layers to improve the quality of learned low-level features~\cite{zhao2017pyramid,zhang2018exfuse}; LabelEnc~\cite{hao2020labelenc} proposes a new label encoding function that maps ground-truth labels to the latent embedding space to add intermediate 
supervisions. 
%
In this work, we automatically select proper intermediate supervisions for generalizable 3D part segmentation networks. 
It can be viewed as an approach that utilizes auxiliary supervisions to improve the network's generalizability. 
\section{Method}

\begin{figure*}[ht]
  \centering
    \includegraphics[width=0.9\textwidth]{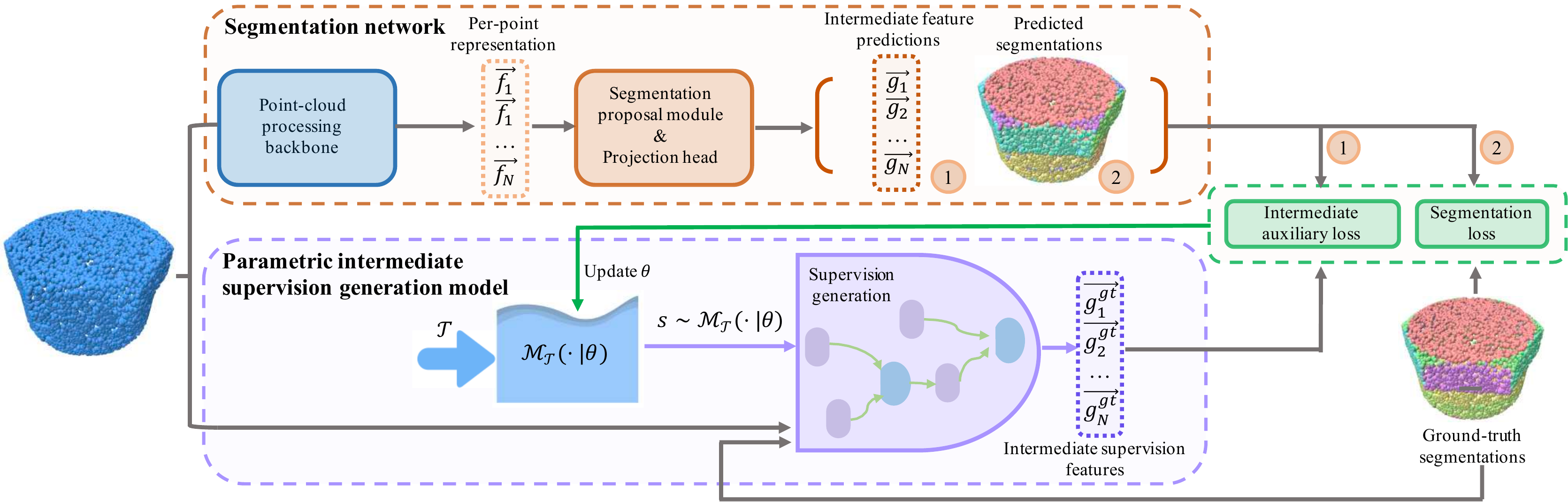}
    \vspace{-5pt}
  \caption{
  An example of applying \gps on an end-to-end segmentation network for the primitive fitting task. 
  \gps builds a learnable parametric supervision model $\mathcal{M}_{\mathcal{T}}(\cdot\vert \theta)$ that can be optimized in order to find proper intermediate supervisions for a generalizable 3D part segmentation network. 
A ``propose, evaluate and update'' strategy is adopted to learn $\theta$.
In each circle, an operation tree is sampled from 
$\mathcal{M}$, 
which is then used to calculate part-aware geometric features for each point. 
The generalization ability of the proposed supervision is evaluated and the 
resulting score is further used to update $\theta$ 
in the updating stage. 
Light purple lines imply the supervision generation process and green for the parameter updating process. 
  }
  \label{fig_overall_pipeline}
  \vspace{-5pt}
\end{figure*}

We present \gpsss, a technique to improve the cross-domain generalizability of a supervised 3D part segmentation network by automatically finding useful intermediate supervisions for the network. 
Then at training time, supervisions found by \gps are introduced in addition to the segmentation supervision 
to alleviate 
the influence of shortcut features~\cite{geirhos2020shortcut} 
and capture real part cues that are invariant across different shapes distributions.
The resulting network can better generalize to shapes from novel distributions without any additional cost at inference time.
Figure~\ref{fig_overall_pipeline} shows an example where \gps is plugged into an end-to-end network for primitive segmentation.

To automatically find such supervisions, 
we design 
a parametric supervision feature space with geometric prior knowledge encoded (\emph{a parametric intermediate supervision model}, Sec.~\ref{sec:3.1}).
To adapt the model for a specific segmentation network, 
we optimize its parameters via 
a ``propose, evaluate, update'' strategy (Sec.~\ref{sec:3.2}). 
After that, a greedy search strategy is utilized to select the optimal supervisions from the 
optimized 
model (Sec.~\ref{sec:3.3}).

\subsection{Parametric Modeling for the Supervision Space}
\label{sec:3.1}

Based on the observation that many previous generalizable 3D part segmentation works guide part feature learning via intermediate supervisions calculated from per-point geometric features and ground-truth part labels,
we assume that being aware of some of such features can help the network learn real part cues.
We propose to automatically search for proper part-aware supervision features for each type of part and add intermediate supervisions to encourage the segmentation network to leverage those features.
%
We then move on to introduce the structural model for supervision features and further 
the supervision space. 

\vspace{-1pt}
\vpara{Structural model for supervision features.}
We define the possible supervision features as the outcome of a tree-structured operation flow (an \emph{operation tree}) taking geometric features and ground-truth part labels as input.  
It is inspired by the calculation process of hand-crafted part-aware geometric features (\eg rotation matrix, cylinder axis~\cite{li2019supervised}, etc.). 

Good supervision features could bridge the gap between input point cloud geometry and the output segmentation labels while avoiding shortcuts. And it is crucial to consider how to use ground-truth labels in order to generate such \textbf{part-aware and geometrically-discriminative} supervision features.
To compute a possible intermediate supervision, instead of treating ground-truth labels as additional input feature vectors, we pre-encode these labels by transferring geometric features of each point to part-aware features. 
We then use such part-aware features as input to an operation tree directly.
An operation tree transforms input features by operators for the output intermediate supervision feature. 
Such operators
include \emph{grouping operators} (\eg sum, SVD\footnote{Singular Value Decompose.}, etc.) that summarize a feature set into a point-level feature, point-level \emph{unary operators} (\eg square, double, etc.), 
and \emph{binary operators} (\eg add, minus, etc.) 
that encourage feature communications to enlarge and diversify
the supervision space.
We further introduce some fixed operator-type combinations named \emph{operation cells} such as a unary operator followed by a grouping operator. 
They are high-frequent operation combinations in the computing process of hand-crafted geometric features~\cite{yi2018deep,li2019supervised}.

An operation tree is then constructed by connecting operation cells.

\begin{figure}[ht]
  \centering
  \vspace{-4pt}
  \resizebox{0.9\linewidth}{!}{%
    \includegraphics[width=0.85\linewidth]{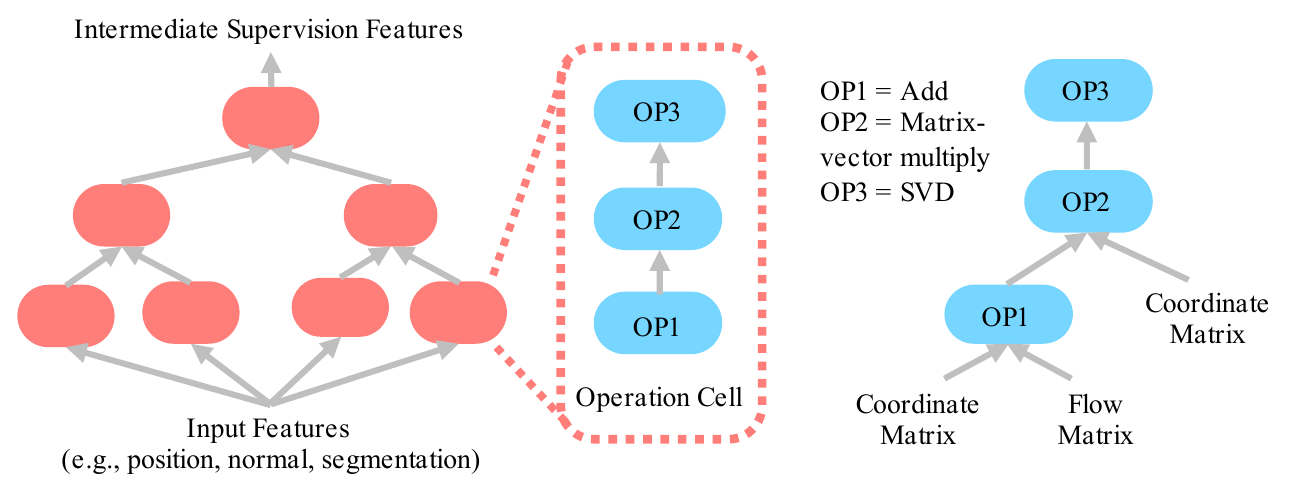}
    }
    \vspace{-4pt}
  \caption{ 
  \footnotesize
  \textbf{Left side: } 
\textbf{Left side:} Supervision feature space. \textbf{Right Side:} An example for the operation tree whose output feature is the rotation matrix calculated by the Orthogonal Procrustes algorithm. Operation cell abstraction is not drawn on it. 
  }
  \label{fig_overall_cell_structure_and_sampling}
\end{figure}

\vspace{-1pt}
\vpara{Supervision feature space.} 
The supervision feature space consists of 
all valid operation trees. 
We set the maximum height of an operation tree to three, thus the supervision feature space is spanned by all possible tree structures and sub-structures of cells in it (Figure~\ref{fig_overall_cell_structure_and_sampling}). 
To measure the generalization benefit of each supervision feature and to optimize the space toward the optimal intermediate supervisions for a segmentation network, 
a parametric distribution model ($\mathcal{M}_{\mathcal{T}}(\cdot \vert \theta)$) is constructed to depict the operation tree space. 
The subscription $\mathcal{T}$ here indicates that it is constructed based on prior knowledge of the task set $\mathcal{T}$, $\theta$ denotes parameters introduced in the distribution model. 
In practice, a marginal distribution is introduced for the grouping operator of the root cell. Then, conditional distributions are introduced for children operators and leaf part-aware features 
conditioned on their parents.

\vspace{-2pt}

\subsection{Learning Task-Conditioned Supervision Distribution}
\label{sec:3.2}
To optimize the distribution parameters $\theta$ for a segmentation network, 
we adopt a 
``propose, evaluate and update'' 
strategy.

\vspace{-4pt}
\vpara{Supervision proposal.} 
Since a supervision feature is modeled by its generation process in our model, an operation tree depicting the generation of a supervision feature is sampled from the parametric model $\mathcal{M}_{\mathcal{T}}(\cdot\vert \theta)$ to generate a supervision feature. 
After that, the input part-aware geometric feature matrices are passed through the sampled operation tree to generate the corresponding supervision feature $s$. 
Then the ground-truth intermediate supervision feature $\vec{g_i}^{gt}$ for each point $i$ is calculated by passing its input features 
through $s$. 

\vspace{-4pt}
\vpara{Supervision evaluation.} 
The cross-domain generalization ability of the proposed supervision feature is estimated under a simulated domain-shift setting. 
Firstly, the training domain is split into different sub-domains. After that, the OOD performance of the sampled supervision feature is estimated by the average generalization error calculated on all of those train-validation splits created by learning one sub-domain out for validation. 
The network is trained together with the proposed supervision $s$ and the segmentation task-related supervision to evaluate the generalization benefit of $s$. 

\vspace{-4pt}
\vpara{Supervision space optimization.} 
The parameter $\theta$ is updated by the probability density value of the generated supervision ($\mathcal{M}_{\mathcal{T}}(s\vert \theta)$) and its estimated generalization score. 
The updating strategy is derived from the REINFORCE~\cite{williams1992simple} algorithm. 
Simply updating parameters of conditional distributions involved in the generation process of $s$ via REINFORCE equals to updating the joint distribution by REINFORCE. 
\subsection{Greedy Supervision Selection}
\label{sec:3.3}

Though we can select a single supervision feature from the supervision model after the optimization (e.g. the feature with the highest probability), we wish to select multiple features to use for better generalizability enhancement. 

To be more specific, 
we adopt a greedy search strategy to select several supervisions (one to three, in our practice) to use further for the segmentation network. 
The greedy search aims to choose an optimal supervision set step-by-step. 
It starts 
from a set containing single supervisions sampled from the optimized supervision space. The generalization ability of such supervision features are then estimated. After that, supervisions with top performance are kept and further used to construct a set containing two supervision pairs. 
Similarly, the performance of supervision combinations are estimated with a new set constructed by coupling top supervisions in each following step. 
Finally, the supervision combination achieved the best performance is then selected to use further. 
\section{Experiments}
We evaluate the effectiveness of the proposed \gps in searching for suitable intermediate supervisions and improving the domain generalization ability of a segmentation network 
on three 3D part segmentation tasks: mobility-based part segmentation~\cite{yi2018deep,hu2017learning}, primitive fitting~\cite{li2019supervised,Yan_2021_ICCV,huang2021primitivenet} and semantic-based part segmentation~\cite{luo2020learning}.

For each task, we set a default segmentation network to evaluate \gps on, which is a point-cloud processing backbone for representation learning followed by a classification-based segmentation module to propose segmentations. 
Two backbones are used in experiments, namely PointNet++~\cite{qi2017pointnetpp} and DGCNN~\cite{wang2019dynamic}. 
We denote such two segmentation networks as ``PointNet++'' and ``DGCNN'' directly for simplicity. 
For each task, we demonstrate that a simple 3D point-cloud segmentation network trained with intermediate supervisions searched by \gps and segmentation task-related supervisions is able to perform better than task-specific models proposed in previous works as well as networks trained by generic domain generalization strategies. 

\begin{figure*}[!tbp]
\centering

  \subfloat{%
    \rotatebox{90}{\scriptsize{~~~~~~~~~~~~~G.T}}
    \includegraphics[width=15.0cm]{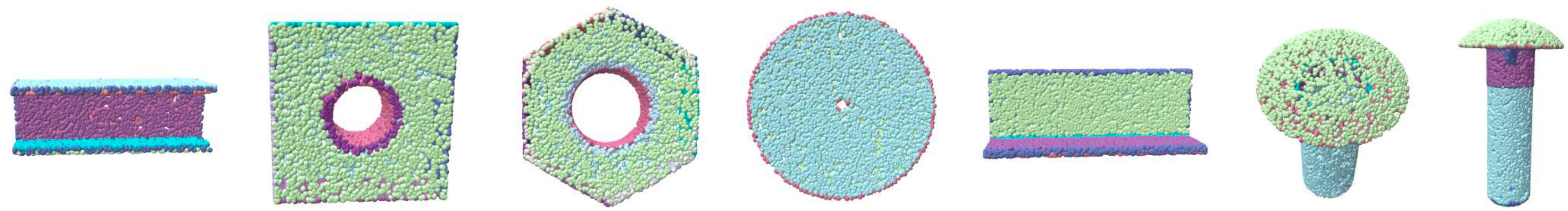}
    }
    
\noindent\rule{\textwidth}{0.5pt}
  \subfloat{%
    \rotatebox{90}{\scriptsize{~~~~~~~~~~~~~Baseline}}
    \includegraphics[width=15.0cm]{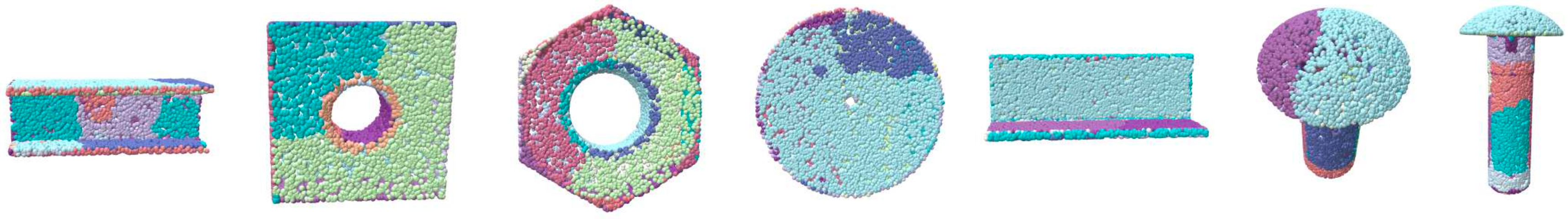}
    }
    
  \subfloat{%
    \rotatebox{90}{\scriptsize{~~~~~~~~~~~~~Ours}}
    \includegraphics[width=15.0cm]{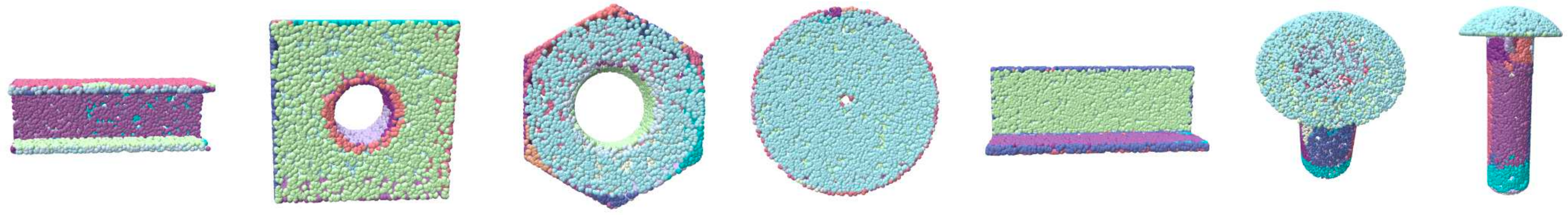}
    }
\vspace{-8pt}
  \caption{Segmentation results visualization for the primitive fitting task. ``Ours'' denotes the model ``\gpsss$_\text{HPNet}$'' and ``Baseline'' denotes the model ``HPNet'' in Table~\ref{tb_prim_fit_seg_eval}.  }
  \label{fig_exp_prim_vis}
 \vspace{-10pt}
\end{figure*}


\subsection{Mobility-based Part Segmentation}


\vspace{-2pt}
\vpara{Datasets.}
Three datasets are used in the task: training dataset, auxiliary training dataset that is only used in the supervision search stage to help simulate domain-shift, and the out-of-domain (OOD) test dataset. 
The training dataset is created from~\cite{yi2016scalable}, containing 15,776 shapes from 16 categories. 
The auxiliary training dataset is created from PartNet~\cite{mo2019partnet}, containing 5,662 shapes from 4 categories different from those used in the training dataset. 
The test dataset used is the same as 
the one used in~\cite{yi2018deep}, which is created from~\cite{hu2017learning}, containing 875 shapes covering 175 objects from 23 categories. 
\vspace{-2pt}
\vpara{Experimental settings.}
We evaluate the effectiveness of \gps on both PointNet++ and DGCNN. 
Metric used for this task is Segmentation Mean IoU (MIoU).
To apply \gps on this task, we add a flow estimation module ahead of the segmentation network.
We compare our method with both task-specific baselines~\cite{yi2018deep,yuan2016space,tzionas2016reconstructing}, and task-agnostic methods~\cite{zhou2021domain,li2018learning}.
Deep Part Induction~\cite{yi2018deep} is trained on the same training dataset using the same settings as 
those 
used for our models. 
All learning-based models use only one forward flow estimation and segmentation proposal pass with no iteration between them.
\vspace{-2pt}
\vpara{Experimental results.}
The performance of \gps and baseline models are summarized in Table~\ref{tb_res_motion_seg_eval}. 
Based on the table, we can make the following observations:
1) \gps can find useful intermediate supervisions that can significantly improve the generalization ability of both PointNet++ and DGCNN 
(\eg 4.9\% absolute MIoU improvement on OOD-test set for PointNet++). 
This can verify the effectiveness of adding intermediate supervisions to boost the generalization ability of segmentation 
networks 
and the ability of \gps to find such supervisions. 
2) \gps can outperform all task-specific models by a large margin, including traditional methods 
(\eg JLinkage clustering) 
and learning based strategies such as Deep Part Induction. 
This may echo the 
proposed 
assumption that hand-crafted supervisions 
may not be optimal ones for the task due to the misalignment between human knowledge and machine understanding.
3) \gps can outperform all task-agnostic strategies when 
using 
the same point-cloud processing backbone. 
A possible reason is that 
geometric prior knowledge which is quite useful to solve the task inherently is
carefully considered in the design of \gps, while task-agnostic strategies are not aware of such prior knowledge. 

\begin{table}[t]
    \centering
    \caption{\footnotesize 
    Experimental results on the mobility-based segmentation task.
    For abbreviations used, In/Out-of-dist. refers to In/Out-of-distribution performance; ``PN++'' denotes ``PointNet++''; ``JLC'' means ``JLinkage clustering''; ``SC'' means ``Spectral Clustering''. Subscriptions indicate the used backbones.
    } 
  \vspace{-8pt}
	\resizebox{0.8\linewidth}{!}{%
    \begin{tabular}{@{\;}c@{\;}|c|c@{\;}}
    \midrule
        \hline
        \specialrule{0em}{1pt}{0pt} 
        Method & In-dist. & Out-of-dist. \\ 
        \cline{1-3} 
        \specialrule{0em}{1pt}{0pt}
      
        \multirow{1}{*}{PointNet++} & 86.4 & 61.0
        \\ \cline{1-3} 
        \specialrule{0em}{1pt}{0pt}
        
        DGCNN~\cite{wang2019dynamic} & {88.3} & 68.1
        \\ \cline{1-3} 
        \specialrule{0em}{1pt}{0pt}
        MixStyle$_{\text{PN++}}$~\cite{zhou2021domain} & 86.6 & 63.4
        \\ \cline{1-3} 
        \specialrule{0em}{1pt}{0pt}
        
        MixStyle$_{\text{DGCNN}}$~\cite{zhou2021domain} & 83.1 & 69.5
        \\ \cline{1-3} 
        \specialrule{0em}{1pt}{0pt}
        
        Meta-learning$_{\text{PN++}}$~\cite{li2018learning} & 71.3 & 63.4
        \\ \cline{1-3} 
        \specialrule{0em}{1pt}{0pt}
        
        Meta-learning$_{\text{DGCNN}}$~\cite{li2018learning} & 76.5 & 69.4
        \\ \cline{1-3} 
        \specialrule{0em}{1pt}{0pt}
        

        JLC~\cite{yuan2016space} & N/A & 67.3
        \\ \cline{1-3} 
        \specialrule{0em}{1pt}{0pt}
        
        SC~\cite{tzionas2016reconstructing} & N/A & 69.4
        \\ \cline{1-3} 
        \specialrule{0em}{1pt}{0pt}
        
        Deep Part Induction~\cite{yi2018deep} & 84.8 & 64.4
        \\ \cline{1-3} 
        \specialrule{0em}{1pt}{0pt}
        
        \gpsss$_{\text{PN++}}$ & {87.2} & 66.5
        \\ \cline{1-3} 
        \specialrule{0em}{1pt}{0pt}
        
        \gpsss$_{\text{DGCNN}}$ & 83.1 & \textbf{73.8}
        \\ \cline{1-3} 
        \specialrule{0em}{1pt}{0pt}

    \end{tabular} 
    }
    \vspace{-14pt}
    \label{tb_res_motion_seg_eval}
\end{table}

\subsection{Primitive Fitting} \label{sec:4.2}


\vspace{-2pt}
\vpara{Dataset.} 
Dataset used in this task is the same as the one used in ~\cite{li2019supervised}. 
However, 
instead of using the provided data splitting method directly,
we re-split the dataset into 4 subsets according to shapes such that it is more suitable to test the cross-domain generalization ability of a model. 
Three subsets out of them, containing 13,528 shapes in total, are used for training, including the supervision search stage and regular training stage for supervision evaluation. 
The left one subset, containing 3,669 shapes, is used for the out-of-domain test. 
For all compared baselines, we test their performance on the re-split dataset for a fair comparison. 

\vspace{-2pt}
\vpara{Experimental settings.} 
We evaluate the effectiveness of \gps on both PointNet++ and DGCNN. 
Metric used for this task is MIoU.
We compare our method with both task-specific baselines~\cite{li2019supervised,Yan_2021_ICCV} and task-agnostic methods~\cite{Mansilla_2021_ICCV,zhou2021domain,li2018learning}.

Among them, HPNet~\cite{Yan_2021_ICCV} adopts a clustering-based two-stage segmentation network for feature learning and segmentation proposal. 
Based on the observation that the high performance achieved by HPNet is largely powered by its clustering-based segmentation module and the synergies between the added supervisions and the clustering module, we conduct the following two experiments to fairly compare \gps with HPNet, since the classification-based segmentation module used in default settings of \gps is not that representative compared with Mean-Shift clustering used in HPNet: 
1) Plug \gps in HPNet's first learning stage to evaluate the effectiveness of \gps in finding useful features that can regularize this learning stage. 
2) Replace the clustering-based segmentation proposal module used in HPNet with a classification-based module to compare the effectiveness of supervisions introduced in HPNet and those searched by \gps under a same segmentation network. 
Resulted models are denoted as ``\gpsss$_\text{HPNet}$'' and ``HPNet*'' respectively in Table~\ref{tb_prim_fit_seg_eval}. 
\vspace{-2pt}
\vpara{Experimental results.}
The performance of \gps and other baseline models are summarized in Table~\ref{tb_prim_fit_seg_eval}. 
We can make the following observations from Table~\ref{tb_prim_fit_seg_eval}: 
1) \gps can find useful intermediate supervisions to improve the generalization ability of a simple segmentation network for the primitive fitting task using either PointNet++ or DGCNN as the backbone, 
similar with that observed for mobility-based part segmentation task. 
2) 
For models using classification-based segmentation modules, 
\gps can boost a simple segmentation network's performance better 
than all task-specific models such as SPFN. 
A possible reason is that the added supervisions, though believed useful by human to help the network learn a correct solution inherently, may not align well with 
the way preferred by machines to solve the task. 
Thus, those hand-crafted supervisions may not be effective enough 
to train a generalizable network. 
Besides, networks may even use shortcut features to optimize those supervisions. 
For models using clustering-based segmentation modules,
\gpsss$_\text{HPNet}$ can achieve better generalization performance than HPNet by adding some intermediate supervisions in its first learning stage. 
It is probably because that the added supervisions searched by \gps can help the network use more real cues to learn per-point features in this stage. 
3) \gps can help better improve the generalization ability of 3D part segmentation networks compared with task-agnostic methods. 

The absolute improvement that \gps adds on HPNet is not as significant as PointNet++ or DGCNN, 
for which we want to emphasize two points as follows: 
1) The high performance achieved by HPNet is largely benefit from the clustering-based segmentation module (\ie 69.6\% MIoU on OOD-test set achieved by HPNet* v.s. 79.5\% achieved by HPNet). 
However, such a segmentation proposal process is quite \textbf{time-consuming}, where about 40 hours are needed to segment all test shapes, while a classification-based one only needs no more than 40 seconds. 
2) 
For classification-based segmentation networks, 
intermediate supervisions searched by \gps are clearly better than that used in HPNet by comparing the performance of \gpsss$_\text{DGCNN}$ with HPNet* (\ie 73.4\% v.s. 69.6\% MIoU on OOD-test set). 
The significant effectiveness of \gps in improving the generalization ability of an end-to-end trained model is of larger practical value. 

\begin{table}[t]
    \centering
    \caption{\footnotesize 
    Experimental results on the primitive fitting task. 
    For abbreviations used, In/Out-of-dist. refers to In/Out-of-distribution performance; ``PN++'' denotes ``PointNet++''. Subscriptions indicate the used backbones; ``GS'' means ``Gradient Surgery''. ``HPNet'' and ``\gpsss$_{\text{HPNet}}$'' use clustering-based segmentation modules. Others use classification-based modules.
    } 
    \vspace{-8pt}
	\resizebox{0.8\linewidth}{!}{%
    \begin{tabular}{@{\;}c@{\;}|c|c@{\;}}
    \midrule
        \hline
        \specialrule{0em}{1pt}{0pt} 
        Method & In-dist. & Out-of-dist. \\ 
        \cline{1-3} 
        \specialrule{0em}{1pt}{0pt}
        
        PointNet++ & 81.5  & 71.6
        \\ \cline{1-3} 
        \specialrule{0em}{1pt}{0pt}
        
        DGCNN & 93.6 & 68.0
        \\ \cline{1-3} 
        \specialrule{0em}{1pt}{0pt}
        
        
        $\text{GS}_{\text{PN++}}$ \cite{Mansilla_2021_ICCV} & 90.7 & 70.3
        \\ \cline{1-3} 
        \specialrule{0em}{1pt}{0pt} 
        
        $\text{GS}_{\text{DGCNN}}$ \cite{Mansilla_2021_ICCV} & 92.6 & 71.6
        \\ \cline{1-3} 
        \specialrule{0em}{1pt}{0pt} 
        
        $\text{Meta Learning}_{\text{PN++}}$ \cite{li2018learning} & 65.0 & 68.5
        \\ \cline{1-3} 
        \specialrule{0em}{1pt}{0pt} 

        $\text{Meta-learning}_{\text{DGCNN}}$ \cite{li2018learning} & 67.1 & 69.3
        \\ \cline{1-3} 
        \specialrule{0em}{1pt}{0pt} 
        
        $\text{MixStyle}_{\text{PN++}}$ \cite{zhou2021domain} & 92.1 & 70.9
        \\ \cline{1-3} 
        \specialrule{0em}{1pt}{0pt} 
        
        $\text{MixStyle}_{\text{DGCNN}}$ \cite{zhou2021domain} & 93.7 & 71.4
        \\ \cline{1-3} 
        \specialrule{0em}{1pt}{0pt} 
        
        
        
        $\text{SPFN}$ \cite{li2019supervised}& {94.4} & 72.3
        \\ \cline{1-3} 
        \specialrule{0em}{1pt}{0pt}
        
        HPNet* \cite{Yan_2021_ICCV} & 93.9 & 69.6
        \\ \cline{1-3} 
        \specialrule{0em}{1pt}{0pt}
        
        HPNet \cite{Yan_2021_ICCV} & N/A & 79.5
        \\ \cline{1-3} 
        \specialrule{0em}{1pt}{0pt}
        
        \gpsss$_{\text{PN++}}$ & 86.3 & 76.5
        \\ \cline{1-3} 
        \specialrule{0em}{1pt}{0pt}
        
        \gpsss$_{\text{DGCNN}}$ & {94.2} & 73.4
        \\ \cline{1-3} 
        \specialrule{0em}{1pt}{0pt}
        
        \gpsss$_{\text{HPNet}}$ & N/A & \textbf{80.4}
        \\ \cline{1-3} 
        \specialrule{0em}{1pt}{0pt}
    \end{tabular} 
    }
    \vspace{-8pt}
    \label{tb_prim_fit_seg_eval}
\end{table}

\subsection{Semantic-based Part Segmentation}

\vspace{-2pt}
\vpara{Dataset.} 
We use the same dataset provided by~\cite{luo2020learning} as well as the train-test data splitting strategy stated in the paper. 

\vspace{-2pt}
\vpara{Experimental Settings.}
%
The evaluation metric used in this task is the Mean Recall value, the same as the one used in~\cite{luo2020learning}. 
Values reported in Table~\ref{tb_res_inst_seg_eval} are the average Mean Recall scores over all test categories. 
We compare \gps with four task-specific methods~\cite{luo2020learning, wang2018sgpn,yi2019gspn,kaick2014shape}, and two task-agnostic approaches~\cite{zhou2021domain,li2018learning}. 


\vpara{Experimental results.}
The performance of \gps and other baseline models are summarized in Table~\ref{tb_res_inst_seg_eval}. 
Different from the above two tasks, where the criterion about segmenting a shape is obvious such as rigid motions, it is not that clear what part cues are useful for this task, meaning not enough prior knowledge to guide a learning-based network's design. 
Thus, the role of ground-truth geometric features is not that important in the previous task-specific designs~\cite{luo2020learning,wang2018sgpn,yi2019gspn}. 
However, their resulting models tend to perform not that well when parsing shapes from novel categories, as shown in Table~\ref{tb_res_inst_seg_eval}, probably due to the influence of shortcut features. 
However, 
our method can help improve the generalization ability of a simple segmentation network to a level comparable to the two-stage learning-based method~\cite{luo2020learning} as well as traditional segmentation methods such as WCSeg~\cite{kaick2014shape}. 
It demonstrates the usefulness of ground-truth part-aware features in providing real cues for a generalizable segmentation network; And also indicates the superiority of \gps to find such useful features for a task where human prior knowledge is not that available or hard to be translated to guide a network's learning process. 

\begin{table}[t]
    \centering
    \caption{\footnotesize 
    Experimental results on the semantic-based part segmentation task.
    For abbreviations used, In/Out-of-dist. refers to In/Out-of-distribution performance. The backbone used in \gps is PointNet++.
    } 
    \vspace{-8pt}
	\resizebox{0.8\linewidth}{!}{%
    \begin{tabular}{@{\;}c@{\;}|c|c@{\;}}
    \midrule
        \hline
        \specialrule{0em}{1pt}{0pt} 
        Method & In-dist. & Out-of-dist. \\ 
        \cline{1-3} 
        \specialrule{0em}{1pt}{0pt}
        
        
        MixStyle\cite{zhou2021domain} & 34.4 & 30.7
        \\ \cline{1-3} 
        \specialrule{0em}{1pt}{0pt}
        
        Meta-learning\cite{li2018learning} & 35.1 & 29.7
        \\ \cline{1-3} 
        \specialrule{0em}{1pt}{0pt}
        
        PartNet-InstSeg\cite{mo2019partnet} & 33.9 & 26.7
        \\ \cline{1-3} 
        \specialrule{0em}{1pt}{0pt}
        
        SGPN\cite{wang2018sgpn} & 25.0 & 20.2
        \\ \cline{1-3} 
        \specialrule{0em}{1pt}{0pt}
        
        GSPN\cite{yi2019gspn} & 23.5 & 28.7
        \\ \cline{1-3} 
        \specialrule{0em}{1pt}{0pt}
        
        Learning to Group\cite{luo2020learning} & 35.2 & 32.0
        \\ \cline{1-3} 
        \specialrule{0em}{1pt}{0pt}
        
        WCSeg\cite{kaick2014shape} & 29.8 & 33.2
        \\ \cline{1-3} 
        \specialrule{0em}{1pt}{0pt}
        
        \gps & {35.7} & \textbf{33.9}
        \\ \cline{1-3} 
        \specialrule{0em}{1pt}{0pt}

    \end{tabular} 
    }
    \vspace{-8pt}
    \label{tb_res_inst_seg_eval}
\end{table}

\subsection{Qualitative Evaluation}
We visualize the segmentation results of  ``\gpsss$_{\text{HPNet}}$'' and ``HPNet'' in Table~\ref{tb_prim_fit_seg_eval} on the Primitive Fitting task for a intuitive comparison and understanding w.r.t. the network's generalization ability on OOD shapes. 
Figure~\ref{fig_exp_prim_vis} shows that \gpsss$_{\text{HPNet}}$ can achieve better segmentation performance on shapes from a novel distribution, i.e., having a relatively large primitive-type distribution gap from that of training shapes. 
A possible reason is that the added intermediate supervisions in the first stage of HPNet can help the model learn per-point features relying more on real part cues, thus less sensitive to the primitive type of each part as well as the primitive-type distribution of the shape. 

\begin{figure}[htbp]
  \centering
    \includegraphics[width=
    0.80
    \linewidth]{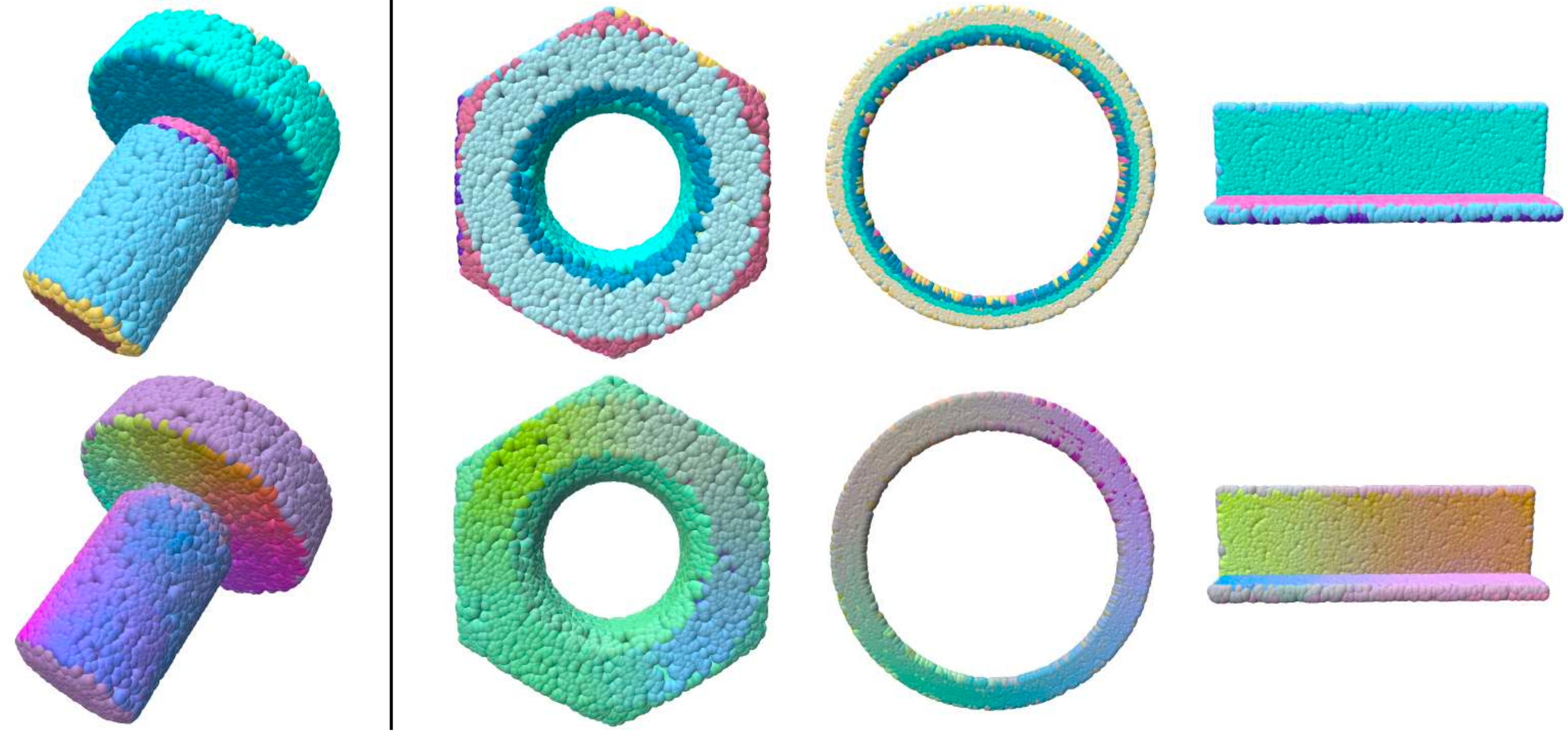}
    \vspace{-7pt}
  \caption{ 
  \footnotesize
  Visualization for one searched intermediate supervision feature from our designed feature space. Upper: Segmentations; Lower: Searched feature (converted to RGB values). The left most shape is the one from the training dataset, while the other three are from the test dataset. 
  }
  \label{fig_vis_intermediate_feature}
\end{figure}

Besides, we draw an intermediate supervision feature searched from our designed supervision space in Figure~\ref{fig_vis_intermediate_feature} for an intuitive understanding towards the property of our searched feature. It can be seen that the searched feature is not only discriminative across different parts, but also changes continuously within one part. Such property may make it be more friendly for the network learning real part cues avoiding shortcut features that only exist in shapes from training distributions. 


\section{Ablation Study}
In our method, we construct an intermediate supervision space and search for useful supervisions from it to increase the generalizability of a segmentation network. In this section, we ablate our method by replacing some crucial designs with other possible alternatives to analyze the effect of those parts. 


\begin{table}[tbp]
    \centering
    \caption{\footnotesize Ablation study w.r.t. reward function design and supervision space design.  For abbreviations used, ``Arch.'' refers to ``Architecture''; ``PN++'' denotes ``PointNet++''; In/Out-of-dist. refers to In/Out-of-distribution performance. 
    } 
    \vspace{-8pt}
	\resizebox{0.80\linewidth}{!}{%
    \begin{tabular}{@{\;}c@{\;}|c|c|c@{\;}}
    \midrule
        \hline
        \specialrule{0em}{1pt}{0pt} 
        Ablation & Arch. & In-dist. & Out-of-dist. \\ 
        \cline{1-4} 
        \specialrule{0em}{1pt}{0pt}
        
        / & \multirow{7}{*}{PN++} & 87.2 & \textbf{66.5}
        \\ \cline{1-1} \cline{3-4}
        \specialrule{0em}{1pt}{0pt}
        
        $-$Cross Validation & ~ & 85.6 & 64.6
        \\ \cline{1-1} \cline{3-4}
        \specialrule{0em}{1pt}{0pt}
        
        $-$Gap & ~ & {90.6} & 64.9
        \\ \cline{1-1} \cline{3-4}
        \specialrule{0em}{1pt}{0pt}
        
        Less operants & ~ & 87.5 & 63.1
        \\ \cline{1-1} \cline{3-4}
        \specialrule{0em}{1pt}{0pt}
        
        Less unary operators & ~ & 86.8 & 64.1
        \\ \cline{1-1} \cline{3-4}
        \specialrule{0em}{1pt}{0pt}
        
        Less binary operators & ~ & 83.0 & 65.1
        \\ \cline{1-1} \cline{3-4}
        \specialrule{0em}{1pt}{0pt}
        
        Less tree height & ~ & 82.1 & 63.5
        \\ \cline{1-4} 
        \specialrule{0em}{1pt}{0pt}
        
        / & \multirow{7}{*}{DGCNN} & 83.1 & \textbf{73.8}
        \\ \cline{1-1} \cline{3-4}
        \specialrule{0em}{1pt}{0pt}
        
        $-$Cross Validation & ~ & 84.5 & 73.1
        \\ \cline{1-1} \cline{3-4}
        \specialrule{0em}{1pt}{0pt}
        
        $-$Gap & ~ & 89.4 & 70.4
        \\ \cline{1-1} \cline{3-4}
        \specialrule{0em}{1pt}{0pt}
        
        Less operants & ~ & {92.1} & 70.6
        \\ \cline{1-1} \cline{3-4}
        \specialrule{0em}{1pt}{0pt}
        
        Less unary operators & ~ & 89.3 & 69.5
        \\ \cline{1-1} \cline{3-4}
        \specialrule{0em}{1pt}{0pt}
        
        Less binary operators & ~ & 89.7 & 71.6
        \\ \cline{1-1} \cline{3-4}
        \specialrule{0em}{1pt}{0pt}
        
        Less tree height & ~ & 89.4 & 71.8
        \\ \cline{1-4} 
        \specialrule{0em}{1pt}{0pt}
    \end{tabular} 
    }
    \vspace{-6pt}
    \label{tb_res_motion_seg_search_cross_val_abl}
\end{table}

\vspace{-1pt}
\vpara{Reward function design.} 
In \gps, we adopt the average generalization gap over all train-validation splits as the reward value to estimate the effectiveness of the selected supervision on improving the model's generalization ability. 
The intuition is that the cross-validating the generalization gap of the searched supervision feature could probably give its generalizability a better estimation. 
In Table~\ref{tb_res_motion_seg_search_cross_val_abl}, we validate the superiority of the designed reward function by replacing it with 1) generalization gap on a single split (denoted as ``$-$Cross validation'' 
), and 2) average 
performance on validation sets 
over all splits (denoted as ``$-$Gap''
). 

\vspace{-1pt}
\vpara{Supervision space design.}
In our method, we use a large supervision feature space to increase the diversity of features in it and also to enable the model a high freedom to select its preferred features by itself at the same time. 
To demonstrate its benefit, 
we downsize the supervision space by downsizing the input part-aware geometric feature set, the unary operator set and the binary operator set respectively and then evaluate the effectiveness of the searched supervisions.
Experimental results prove the superiority of using a large supervision space (see Table~\ref{tb_res_motion_seg_search_cross_val_abl}).
%
%

\begin{figure}[htbp]
    \flushleft
    \includegraphics[width=
    0.90
    \linewidth]{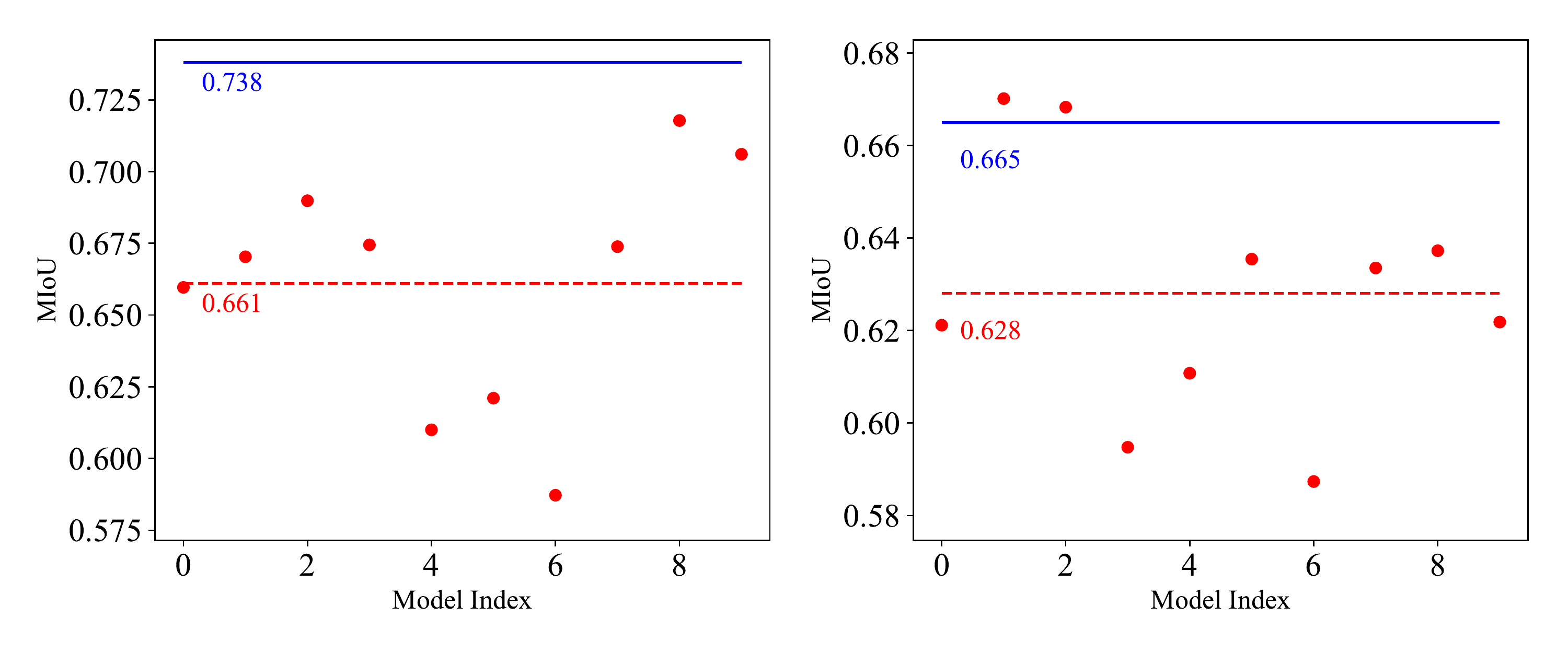}
    \vspace{-7pt}
  \caption{ 
  \footnotesize
  Ablation study w.r.t. supervision selection strategy. 
  Ten supervision features are randomly selected from the optimized distributions and evaluated. 
  Left for PointNet++ and right for DGCNN.
  }
  \label{fig_abl_random_pnpp_dgcnn}
\end{figure}

\vspace{-1pt}
\vpara{Supervision selection strategy.}
In our method, we use a greedy search-like strategy to select supervision features for future use based on the intuition that such a strategy can help us find a good supervision feature combination from the optimized space. 
To demonstrate the effectiveness of the greedy supervision selection strategy used in our method, we try to ablate it and consider the following two alternatives: 
1) Select top $K (1\le K \le 3)$ features from the 
supervision feature set 
ordered 
by the probability density value.
Results 
are summarized in Table~\ref{tb_res_motion_seg_search_sup_sel_stra}. 
2) Choose features from the optimized distributions randomly. 
Ten random features are sampled and their results are summarized in
Figure~\ref{fig_abl_random_pnpp_dgcnn}.

\begin{table}[htbp]
    \centering
    \caption{\footnotesize
    Ablation study w.r.t. the supervision selection strategy. 
    For abbreviations used, ``Arch.'' refers to ``Architecture''; ``PN++'' denotes ``PointNet++''; In/Out-of-dist. refers to In/Out-of-distribution performance. 
    } 
    \vspace{-10pt}
	\resizebox{0.70\linewidth}{!}{%
    \begin{tabular}{@{\;}c@{\;}|c|c|c@{\;}}
    \midrule
        \hline
        \specialrule{0em}{1pt}{0pt} 
        Ablation & Arch. & In-dist. & Out-of-dist. \\ 
        \cline{1-4} 
        \specialrule{0em}{1pt}{0pt}
        
        / 
        & \multirow{4}{*}{PN++} & {87.2} & \textbf{66.5}
        \\ \cline{1-1} \cline{3-4}
        \specialrule{0em}{1pt}{0pt}
        
        Top1 & ~  & 87.1 & 65.6
        \\ \cline{1-1} \cline{3-4}
        \specialrule{0em}{1pt}{0pt}
      
        Top2 & ~  & 81.5 & 64.5
        \\ \cline{1-1} \cline{3-4}
        \specialrule{0em}{1pt}{0pt}
        
        Top3 & ~ &  85.8 & 62.8
        \\ \cline{1-4}
        \specialrule{0em}{1pt}{0pt}
        
        / 
        & \multirow{4}{*}{DGCNN} & 83.1 & \textbf{73.8}
        \\ \cline{1-1} \cline{3-4}
        \specialrule{0em}{1pt}{0pt}
        
        Top1 & ~ & {92.1} & 70.1
        \\ \cline{1-1} \cline{3-4}
        \specialrule{0em}{1pt}{0pt}
      
        Top2 & ~  & 89.8 & 67.8
        \\ \cline{1-1} \cline{3-4}
        \specialrule{0em}{1pt}{0pt}
        
        Top3 & ~ &  90.7 & 71.6
        \\ \cline{1-4}
        \specialrule{0em}{1pt}{0pt}
        
    \end{tabular} 
    }
    \vspace{-10pt}
    \label{tb_res_motion_seg_search_sup_sel_stra}
\end{table}




\section{Discussion and Conclusion}


In this paper, we propose to automatically search for proper intermediate supervisions for generalizable 3D part segmentation networks. 

Although experimental results 
can prove the effectiveness and versatility of \gps to some extend, there are still many directions worth exploring by the community in the future: 
1) Search intermediate supervisions and the backbone's structure at the same time. Synergies between backbone architecture and intermediate supervisions may be found since different backbones may prefer different features. 
2) How to add intermediate supervisions. Directly predicting a ground-truth feature may not be suitable for each feature. Moreover, synergies between supervisions features and how to supervise the network to learn such features may be discovered. 
{\small
\bibliographystyle{ieee_fullname}
\bibliography{main}
}

\clearpage



\appendix

\section{Overview}
The 
appendix 
aims at providing more details on the method design (Sec.~\ref{sec_method_details}), experimental settings (Sec.~\ref{sec_exp_details}), and more experimental results (Sec.~\ref{sec_additional_res}).

\section{Method Details} \label{sec_method_details}
In this section, we give some explanations for some details of the proposed method that are not stated in the main paper.

\subsection{Supervision Feature Structure}
This section 
gives more details
about designed structural model for generating ground-truth part-aware geometric features w.r.t. further explanations for each component, operation cells and the rationality of the design.
 
\vpara{Input features.}
Input features for each point are ordered part-aware feature sets formed by 
different kinds of geometric features from points having the same part labels with the target point.
A maximum radius is set between the sampled points and the target point considering the limited receptive field of some point-cloud processing backbones such as DGCNN. This value is set to 0.5 in our implementation.

The full set of ordered input feature sets are formed by first sampling 
such part-aware feature sets from input geometric features such as coordinate vectors, then expanding the resulting sets by adding feature sets calculated by simple point-level operations among them to it.
For example, suppose each point $i$ has two different input geometric features: the coordinate vector $\vec{p}_i$ and the normal vector $\vec{n}_i$.
Their ordered respective part-aware feature sets for point $i$, constructed by sampling same kind of features from points having the same part labels with it, are denoted as $\mathcal{P}_i$ and $\mathcal{N}_i$ respectively.
In the first step, we can get $\{\mathcal{P}_i, \mathcal{N}_i\}$ for point $i$.
Then after expanding the set, we get $\{ \mathcal{P}_i, \mathcal{N}_i, \mathcal{P}_i - \mathcal{N}_i... \}$, where $\mathcal{P}_i - \mathcal{N}_i$ refers to a feature set formed by element-wise minus between $\mathcal{P}_i$ and $\mathcal{N}_i$.
An ordered part-aware geometric input feature set such as $\mathcal{P}_i$ is referred as an ``operant'' in the operation tree.
In practice, an ordered feature set is formed to a matrix with each of its line a feature vector of a point, thus $P_i$ for $\mathcal{P}_i$ and $N_i$ for $\mathcal{N}_i$.
If each point has two different geometric features, the full set of input part-aware geometric matrices contains 7 elements: $\{ P_i,N_i,P_i\cdot N_i,P_i+N_i,P_i-N_i,N_i-P_i,\text{cross\_product}(N_i,P_i) \}$.
If each point has one geometric feature, the full set of input part-aware geometric matrices contains 4 elements: $\{P_i,2P_i,P_i^2,-P_i \}$.

\vpara{Operators.}
Operators are introduced to transform the input features step by step for the output intermediate supervision feature. 
Such operators include grouping operators, point-level unary operators, and point-level binary operators, as summarized in Table~\ref{tb_method_tree_gen_ops}.
Then the grouping operator set, the unary operator set, and the binary operator set are further referred as $\mathcal{G}$, $\mathcal{U}$, and $\mathcal{B}$ respectively.

Some operators may seem confusing, for which we give them some brief explanations as follows: 
\begin{itemize}
    \item Orthogonalize: Given a matrix $M$, first calculate its signular vectors and singular values by $U, S, V^T = \text{SVD}(M)$, then take the matrix-vector multiplication between two singular vector matrices for the result: $\text{Orthogonalize}(M) = UV^T$.
    \item Cartesian Product: Given two feature sets from one point, $\mathcal{S}_1$ and $\mathcal{S}_2$, calculate the cartesian product between them by $\mathcal{S}_1 \times \mathcal{S}_2 = \{ (s_i, s_j) | s_i\in \mathcal{S}_1, s_j \in \mathcal{S}_2 \}$. 
\end{itemize}

\vpara{Operation cells.}
We introduce several different kinds of operation cells with fixed operator combinations to encourage some fixed operation sequences. 
Details of such operator combinations for operation cells in different levels are listed as follows: 
\begin{itemize}
    \item Level-3 cells (Top level cells): a grouping operator followed by a unary operator. 
    \item Level-2 cells: a grouping operator followed by a uanry operator.
    \item Level-1 cells: an operant followed by a uanry operator.
\end{itemize}



\begin{table}[t]
    \centering
    \caption{\footnotesize 
    Candidates of each kind of feature transformation operator. 
    ``SVD'' denotes ``Singular Value Decompse'', ``Identity'' refers to no operations, where the input ordered feature set (matrix) is not transformed by any unary operator.
    } 
    \vspace{-8pt}
		\resizebox{0.9\linewidth}{!}{%
    \begin{tabular}{@{\;}c@{\;}|c@{\;}}
    \midrule
        \hline
        \specialrule{0em}{1pt}{0pt} 
        Operators & Choices \\ 
        \cline{1-2} 
        \specialrule{0em}{1pt}{0pt}
        Grouping operators & Sum, Average, Maximum, SVD
        \\ \cline{1-2} 
        \specialrule{0em}{1pt}{0pt}
        
        Binary operators & \makecell[c]{Add, Minus, Multiply, \\ Cross Product, Cartesian Product, \\  Matrix-vector Product} 
        \\ \cline{1-2} 
        \specialrule{0em}{1pt}{0pt}
        
        Unary operators &  \makecell[c]{Identity, Square, Double,  \\ Negative, Orthogonalize, Inverse}
        \\ \cline{1-2} 
        \specialrule{0em}{1pt}{0pt}
    \end{tabular} 
    }
    \label{tb_method_tree_gen_ops}
        \vspace{-8pt}

\end{table} 

\vpara{Justification for the designed operation cells.}
In the design of operation cells, we set some fixed operator sequences for them such as a grouping operator followed by a unary operator.
The design of the structure of operation cells is heuristic but also reasonable.
For example, a unary operator following a grouping operator can further transform the grouped feature, extracting part-level information from it.
Such calculation routines are common in the calculation process of some features such as the rotation matrix $R$. 
To be more specific, several steps in the calculation process of $R$ contain a combination of ``Centralize'' and ``Sum'', where the former is a unary operator and the later is a grouping operator.
This is only one possibility we explored in our practice, we believe there could be many other strategies to design operation cells (e.g. other computing routines) or just constructing the operation tree without introducing operation cells.
How to design suitable operation cells, or how to add proper constraints for operation combinations in the operation space is interesting and worth exploring.

\vpara{Connections with hand-crafted geometric features.}
The constructed supervision feature space contains many hand-crafted geometric features used in previous works.
Some examples are summarized in Table~\ref{tb_feat_calculation}.
The calculation process uses matrix forms of the ordered geometric feature sets.

\begin{table}[t]
    \centering
    \caption{\footnotesize Examples of hand-crafted ground-truth features used in previous works. Matrices in the table are part-level matrices formed by corresponding features from points in a part. For notations used, $F$ is the flow matrix, $P$ is the coordinate matrix, $N$ is the normal matrix. $\overline{M}$ denotes the centralized matrix of $M$. Note that the geometric feature listed in the left column of the table may not be the exact feature output by the calculation process listed in its corresponding right cell, but a part of the output feature such as a row vector of the matrix. 
    } 
    \vspace{-8pt}
		\resizebox{0.9\linewidth}{!}{%
    \begin{tabular}{@{\;}c@{\;}|c@{\;}}
    \midrule
        \hline
        \specialrule{0em}{1pt}{0pt} 
        Feature & Calculation process \\ 
        \cline{1-2} 
        \specialrule{0em}{1pt}{0pt}
      
        Rotation matrix & \makecell[c]{ $\text{orth}(\text{sum}(\text{cartesian}(P + F, P)))$ } 
        \\ \cline{1-2} 
        \specialrule{0em}{1pt}{0pt}
        
        Cone apex & $\text{sum}(\text{cartesian}(N^{+T}, \text{rowsum}(N\cdot P)))$ 
        \\ \cline{1-2} 
        \specialrule{0em}{1pt}{0pt}
        
        Cylinder axis & $\text{svd}(\overline{N})$ 
        \\ \cline{1-2} 
        \specialrule{0em}{1pt}{0pt}
        
        Sphere center & $\text{sum}(\text{cartesian}((-2\overline{P})^{+T}, \text{rowsum}(P^2)))$ 
        \\ \cline{1-2} 
        \specialrule{0em}{1pt}{0pt}
        
        Sphere radius & $\text{sqrt}(\text{mean}(\text{P} - \vec{c})^2)$ 
        \\ \cline{1-2} 
        \specialrule{0em}{1pt}{0pt}
        
        Plane normal & $\text{svd}(\overline{P})$ 
        \\  \cline{1-2} 
        \specialrule{0em}{1pt}{0pt}

    \end{tabular} 
    }
    \vspace{-6pt}
    \label{tb_feat_calculation}
\end{table} 

\subsection{Supervision Feature Distribution Space}
This section aims for some further explanations of the constructed supervision feature distribution space w.r.t. how we decompose it into a tree-structured distribution space and the conditional sampling process performed on the distribution tree.

\vpara{Decomposed tree-structured distribution set.}
Instead of using one single distribution, we decompose the total/joint distribution into a tree-structured distributions set to better model the generation process of the operation tree.
Specifically, we use a distribution cell to depict the space of an operation cell, where the operator sequence in the operation cell is generated by a set of conditional operator distributions.
For instance, a distribution cell for an operation cell with the ordered operator sequence [a grouping operator, a unary operator] is composed of a grouping operator distribution and $\vert \mathcal{G}\vert$ unary operator distributions.
Each possible grouping operator has its own unary operator distribution.
Thus, the distribution cell for this specific operation cell can be organized into a distribution tree with the grouping operator distribution as the top distribution and a unary operator distribution attached to each element in the grouping operator.

To model possible connections between cells from adjacent two levels, we introduce a connection distribution for each possible operator sequence of the upper-level cell to depict each possible connection and the operator that should be used for connection.
For example, a Level-3 cell can be connected with two Level-1 cells with each kind of binary operator, or a Level-2 cell and a Level-1 cell with each kind of binary operator, or a Level-1 cell and a Level-2 cell with each kind of binary operator, or a single Level-1 cell with no binary operator.
Thus, the total number of possible connections is $3\vert \mathcal{B}\vert + 1$.
A distribution containing elements of this number is introduced for each possible operator sequence of the upper-level cell.


\vpara{Conditional operation tree sampling.}
After the distribution space has been constructed, a conditional sampling process is adopted to sample an operation tree from the space.
Specifically, to sample an operation cell from its respective distribution cell, its top operator is sampled at first.
Then, the distribution for the following level operator of this top operator is chosen to continue the sampling process.
Specifically, the sampling process for sampling an operation sequence [$\text{op}_1, \text{op}_2, ..., \text{op}_k$] is conducted by: Sample $\text{op}_k$, conditioned on $\text{op}_k$ and sample $\text{op}_{k-1}$, ..., conditioned on $\text{op}_k,\text{op}_{k-1},...,\text{op}_{2}$ and sample $\text{op}_1$.
The probability density value of the sequence is then calculated by:
\begin{align}
    p([\text{op}_1, \text{op}_2, ..., \text{op}_k]) = \\ p(\text{op}_k)p(\text{op}_{k-1}|\text{op}_k)...p(\text{op}_1|[\text{op}_k,...,\text{op}_2]).
\end{align}

Similarly, to sample an operation tree, the top operation cell is first sampled. 
Then, a connection is sampled conditioned on the sampled operation cell structure.
The sampled connection determines what distribution cells to continue the sampling process and what binary operator to use for connection if needed.
Thus, the operation tree can be sampled by such a top-down manner.



\subsection{Supervision Feature Distribution Learning}
In this section, we talk about how to learn parameters for distributions in the supervision feature space, including the supervision feature sampling process, its cross-domain generalization ability evaluation and the distribution updating process.

\vpara{Supervision feature sampling.}
A supervision feature is generated by first sampling an operation tree from the constructed distribution space and then pass input part-aware geometric features of each point through the operation tree to get its calculated part-aware geometric feature.
The operation tree is sampled by the conditional sampling process from the distribution space as state above.

\vpara{Cross-validation setting.}
Each selected supervision is put under a cross-validation process to estimate its cross-domain generalization ability.
To cross-validate the generalization ability of the selected supervision, we first split the train dataset into $K$ subsets with a relatively large distribution shift across them.
Then, the selected supervision is tested on each $K-1:1$ train-validation fold.
The average value between the metric on the validation set and the train set is taken as the estimated score for its generalization ability.
In the supervision search process, the generalization score is further taken as the reward for the selected supervision and used for the following supervision distribution space update process.
The cross-validation procedure for generalization gap evaluation is summarized in Algorithm~\ref{algo_cross_val}.

\vpara{Supervision distribution space optimization.}
We optimize distributions in each layer related with the sampling process of the selected supervision via the REINFORCE algorithm.
Optimizing those conditional distributions via REINFORCE results optimizing the overall/joint distribution via REINFORCE.
The optimization strategy is derived from the REINFORCE algorithm, where each parameter $w_{i}$ that counts in the sampling process is updated by minusing its corresponding $\Delta w_{i} = \alpha_{i}(r - b_{i})e_i$, where $e_{i} = \partial \ln g / \partial w_{i}$, $g$ is the probability density function. 
Thus, $e_{i}$ for parameter $w_{i}$ of a distribution can be derived as: 

\begin{align}
    e_i &= \frac{\partial g}{\partial w_i} \\ 
    &= \frac{\partial \ln( g(\mathcal{H})g(v| \mathcal{H})g(\mathcal{C}\setminus \{v\}| \mathcal{H})g(\mathcal{L}|\mathcal{H},\mathcal{C}) )}{\partial w_i} \\
    &= \frac{\partial \ln(g(\mathcal{H})  + \partial \ln(g(v| \mathcal{H})) + \partial  \ln(g(\mathcal{L} | \mathcal{H},v)) )} {\partial w_i} \\
    &= \frac{\partial \ln(g(v| \mathcal{H})}{\partial w_i},
\end{align}

where $g(\mathcal{H})$ is the probability density of values sampled in upper layers, $g(\mathcal{L} | \mathcal{H}, \mathcal{C})$ is the probability density of values sampled in lower layers conditioned on values sampled in current layers $\mathcal{C}$ and those sampled in upper layers $\mathcal{H}$, which are not relevant with the sampling distribution in the current layer resulting value $v$. 
$\mathcal{H}$ denotes the set of operators in upper layers of the operation tree, $\mathcal{C}$ means the set of operators in the current layer of the operation tree, while $\mathcal{L}$ refers to the set of operators in lower layers.
The conditional distribution for each sample $g(v| \mathcal{H})$ is a multinomial distribution, in our implementation, which can be simply calculated by the class function ``log\_prob'' of the corresponding PyTorch Distribution module.

\subsection{Greedy Supervision Feature Selection}
This section aims for some further explanations of the unstated details of the greedy feature selection process.

After we have got the optimized supervision distribution set, we greedily select suitable supervisions from the optimized space.
At most three supervisions are selected from the optimized space.
To complete the supervision selection process, we first select $K$ supervisions from the optimized space by randomly sampling 
from the optimized supervision space.
Then the performance of such $K$ supervisions are evaluated under a simulated domain-shift setting.
From the ranked set, top 2 supervisions are selected to combine with top $K / 2$ supervisions and form the second supervision set.
Supervision combinations in the second supervision set are further evaluated and ranked.
Then, top 3 supervision combinations containing 2 supervisions are selected to combine with top $K / 3$ single supervisions from the first supervision set.
The resulting combination set is further evaluated and ranked.
The supervision combination that achieves the best estimated performance is selected for further evaluation.
The performance of the selected supervision combination is regarded as the performance of the optimized supervision distribution space.
The training stage of the supervision evaluation process is referred as the ``regular training stage''.
This greedy selection process is also summarized in Algorithm~\ref{algo_greedy_sel}.

\begin{algorithm}[htbp]
\small 
\caption{Cross\_Val. ``TrainValidation($\cdot, \cdot, \cdot, \cdot$)'' takes a model, the train dataset, the validation dataset and the number of training epochs $n$ as input, trains the model for $n$ epochs and returns the gap between the performance of the model achieved on the validation set and the training set at the best validation epoch.
}
\label{algo_cross_val}
\footnotesize
    \begin{algorithmic}[1]
        \Require
            The model $\mathcal{M}$; Split train set $\mathcal{S}_{sp} = \{ \mathcal{S}_1, \mathcal{S}_2, ..., \mathcal{S}_k \}$; Intermediate supervision feature $t$; Epochs for training $n$.
        \Ensure
            Estimated generalization score which is actually the average generalization gap across all train-validation splits.
        
        \State $\text{scores} \leftarrow [~]$
        \For{$i = 1$ to $\vert\mathcal{S}_{sp}\vert$}
            \State $\mathcal{S}_{tr}\leftarrow \mathcal{S}_{sp} \setminus \{ \mathcal{S}_i \}$
            \State $\mathcal{S}_{val}\leftarrow \{ \mathcal{S}_i \}$
            \State $s_i \leftarrow \text{TrainValidation}(\mathcal{M}, \mathcal{S}_{tr}, \mathcal{S}_{val}, n)$
            \State $\text{scores}.\text{append}(s_i)$
        \EndFor
        
        \State $\overline{s} \leftarrow \text{mean}(\text{scores})$ \\
        \Return $\overline{s}$
       
    \end{algorithmic}
\end{algorithm}

\begin{algorithm}[H]
\small 
\caption{SortByCrossVal. 
}
\label{algo_sorted_by_cross_val}
\footnotesize
    \begin{algorithmic}[1]
        \Require
            The model $\mathcal{M}$; Split train set $\mathcal{S}_{sp} = \{ \mathcal{S}_1, \mathcal{S}_2, ..., \mathcal{S}_k \}$; A set of intermediate supervision features $\mathcal{T} = \{t_1, t_2, ..., t_K\}$.
        \Ensure
            Sorted intermediate supervision features $\mathcal{T}'$.
        
        \For{$i = 1$ to $\vert\mathcal{T}\vert$}
            \State $s \leftarrow \text{Cross\_Val}(\mathcal{M}, \mathcal{T}_1[i], \mathcal{S}_{sp})$
            \State $\mathcal{T}_1[i] \leftarrow (\mathcal{T}_1[i], s)$
        \EndFor
        
        \State $\mathcal{T}' \leftarrow \text{sorted}(\mathcal{T})$ \\
        \Return $\mathcal{T}'$
       
    \end{algorithmic}
\end{algorithm}

\begin{algorithm}[H]
\small 
\caption{GreedySupervisionSelection. 
}
\label{algo_greedy_sel}
\footnotesize
    \begin{algorithmic}[1]
        \Require
            The model $\mathcal{M}$; Split train set $\mathcal{S}_{sp} = \{ \mathcal{S}_1, \mathcal{S}_2, ..., \mathcal{S}_k \}$; A set of sampled intermediate supervision features $\mathcal{T} = \{t_1, t_2, ..., t_K\}$.
        \Ensure
            Selected intermediate supervision feature set $\mathcal{T}_s = \{t_{s1}, ..., t_{sk} \}, 1\le k \le 3$. 
        
        \State $\mathcal{T}_1 \leftarrow \mathcal{T} $
        \State $\mathcal{T}_1 \leftarrow \text{SortByCrossVal}(\mathcal{M}, \mathcal{T}_1, \mathcal{S}_{sp})$
        \State $\mathcal{T}_2 \leftarrow \mathcal{T}_1[1: 2] \times \mathcal{T}_1[1: \vert \mathcal{T}_1 \vert / 2]$
        \State $\mathcal{T}_2 \leftarrow \text{SortByCrossVal}(\mathcal{M}, \mathcal{T}_2, \mathcal{S}_{sp})$
        \State $\mathcal{T}_3 \leftarrow \mathcal{T}_2[1: 3] \times \mathcal{T}_1[1: \vert \mathcal{T}_1\vert / 3]$
        \State $\mathcal{T}_3 \leftarrow \text{SortByCrossVal}(\mathcal{M}, \mathcal{T}_3, \mathcal{S}_{sp})$
        \State $\mathcal{T}_s \leftarrow \text{sorted}(\{\mathcal{T}_1[1], \mathcal{T}_2[1], \mathcal{T}_3[1] \})[1]$ \\ 
        \Return $\mathcal{T}_s$
       
    \end{algorithmic}
\end{algorithm}










\section{Experimental Details} \label{sec_exp_details}
In this section, we provide some detailed information about experimental settings that are not stated in the main paper, including datasets used in each segmentation task, detailed settings for each stage, implementation for baseline models, etc.

\subsection{Dataset}

\vpara{Mobility-based part segmentation.}
For the training dataset and auxiliary training dataset, we first infer part mobility information for parts in each shape. 
Then, during the training process, we generate shape pairs for training based on such part mobility information.
For the training dataset, which is created from~\cite{yi2016scalable}, containing 15,776 shapes from 16 categories, we first infer mobility information for parts in a shape heuristically based on their semantic labels such that the generated mobility information for the part can align better with real scenarios. 
Specifically, some heuristic constraints are added to infer mobility information for each part such as the chair back can rotate around the chair seat but not chair legs.
For the auxiliary dataset created from PartNet~\cite{mo2019partnet}, the mobility information for parts in a shape is also inferred heuristically where only some general motion rules are added, which means that parts of different semantic labels share the same motion rules. 
The test dataset used in our work is the same as the one used in~\cite{yi2018deep}, which is created from~\cite{hu2017learning}.

\vpara{Primitive fitting.} 
We use the same dataset as the one provided by~\cite{li2019supervised}, but adopt a different data splitting strategy such that it is more suitable to test a model's cross-domain generalization ability. 
Specifically, we split shapes via their primitive-type distributions. 
Primitive-type distribution reveals the ratio of each primitive-type in the shape calculated based on number of points belonged to each type of primitive.
We first cluster all shapes into 7 clusters by K-Means++~\cite{arthur2006k}, then merge them into 4 subsets with a relatively large distribution gap across them. 
``Train\_1'', ``Train\_2'', and ``Train\_3'' are all used in the supervision search process and regular training process.
In the supervision search process, such three train splits serve for distribution-shift simulation to cross-validate the cross-domain generalization ability of the selected supervision.
In the regular training process, shapes in those three splits are merged together and further split into train-validation datasets via a ratio 9:1.
The validation set is used for model selection.

\vpara{Semantic-based part segmentation.}
The dataset we use is the same as the one used in~\cite{luo2020learning}. 
We only use the finest segmentation level for training and evaluation other than using all available levels as does in~\cite{luo2020learning}.

\subsection{Experimental Settings}

\vpara{Supervision search.} 
For mobility-based part segmentation, 100 automatic search epochs are conducted. 
For primitive fitting and semantic-based part segmentation, 30 automatic search epochs are conducted. 
For all three tasks, 4 supervisions sampled and evaluated in each epoch.
The supervision distribution space is optimized in each epoch. 
AdaM optimizer is used for segmentation networks for all three tasks, with $\beta=(0.9, 0.999),\epsilon=10^{-8}$, and weight decay ratio set to $10^{-4}$. 
Batch size is set to 36 for mobility-based part segmentation task, 2 for primitive fitting and semantic-based part segmentation using DGCNN, 8 those two tasks using PointNet++.

\vpara{Regular training.} 
We use AdaM optimizer for segmentation networks for all three tasks, with $\beta=(0.9, 0.999),\epsilon=10^{-8}$, and weight decay ratio set to $10^{-4}$. 
For mobility-based part segmentation networks, 400 epochs are performed with the epoch that the model achieves the best validation performance is take for inference. 
For primitive fitting and semantic-based part segmentation tasks, 200 epochs are performed using the same model selection strategy as the one for the mobility-based part segmentation networks.
When using clustering-based segmentation module, 100 training epochs are conducted with the model achieves the lowest validation contrastive-style loss further used for inference.
As for the cross-validation strategy used for estimating the effectiveness of the selected supervision in improving the network's domain generalization ability, two training datasets are used for cross-validation for the mobility-based part segmentation task, namely the training dataset and the auxiliary training dataset.
While three training datasets are used for both primitive fitting and the semantic-based part segmentation.
Three clusters out of four clusters are used for cross-validating in primitive fitting task.
Three categories, including ``Chair'', ``Lamp'', and ``StorageFurniture'', are used for cross-validating semantic-based part segmentation.

As for the whole cross-validating process, we train the network on each train-validation split fold for one single epoch with intermediate supervisions added based on the selected supervision feature and the segmentation task related supervision optimized simultaneously.
After that, the average metric gap across all train-validation splits is taken as the estimated generalization score for the selected supervision feature, which is also used as the reward score for further supervision feature distribution space update.

\vpara{Ablation Study.}
For details of different ablated versions w.r.t. the supervision space design. ``Less operants'' denotes using a smaller input feature candidate set that is not expanded. For a set of input part-level matrix cadidates we used in the full regular searching process where the full version of the input feature set that is expanded from input features, only the set containing part and geometry-aware matrices formed from input features directly is used in this version. 
More specifically, in ``Less operants'' setting, we use \{ $P_i, N_i$\} as the input feature set, while the full version contains 7 matrices.
``Less unary operators'' denotes the version where only a subset of unary operators is used as the unary operator set.
Compared with the full version listed in Table~\ref{tb_method_tree_gen_ops}, the one used in the ablated version is \{Identity, Square, Double, Negative\}.
``Less binary operators'' refers to the version where only a subset of binary operators is used as the binary operator set.
Compared with the full version listed in Table~\ref{tb_method_tree_gen_ops}, the one used in the ablated version is \{Add, Minus, Multiply\}.
``Less tree height'' means the maximum height of the operation tree, which is measured by the number of the connected operation cells. 
Compared with the one used in the full version that is set to 3, the maximum tree height in the ablated version is set to 2.

\vpara{Models for comparison with HPNet.}
In the primitive fitting task, we develop two models to compare with HPNet fairly, considering the clustering-based network architecture used in HPNet that is different from the classification-based segmentation module used in our default setting to evaluate \gps.
Two models are designed by 1) replacing the clustering-based segmentation module used in HPNet with a classification-based segmentation module, denoted as ``HPNet*''; 2) plug \gps in the first learning stage of HPNet to search for useful intermediate supervisions that can help the network learn  representations using more invariant features and avoid using shortcut features~\cite{geirhos2020shortcut}, denoted as ``\gpsss$_\text{HPNet}$''.
Following are some detailed settings for such two models.
For HPNet*, the network is formed by replacing the clustering-based segmentation module with a classification-based segmentation module with supervisions added on the first learning stage kept.
The model is trained and evaluated using the same setting as for our own model.
That is, train the model for 200 epochs and select the best validation epoch for further evaluation.
For \gpsss$_\text{HPNet}$, we also adopt a two-stage training procedure. 
In the first stage, \gps is applied to search for useful intermediate supervisions using the gap of the contrastive-style loss between the training dataset and the validation dataset across all train-validation splits as the reward.
After the supervision distributions have been optimized, we greedily select what supervisions to use and plug them in the first learning stage of HPNet by optimizing such losses and other losses introduced in HPNet simultaneously.
The resulting model optimized in the first learning stage is then taken for further evaluation using the clustering-based segmentation module.

\vpara{Point-cloud processing backbones.}
PointNet++ used in our default setting is the same one as that used in SPFN~\cite{li2019supervised}.
DGCNN used in the default setting is the same one as that used in HPNet~\cite{Yan_2021_ICCV} for representation learning.

\vpara{Input features.}
For semantic-based part segmentation task where per-point normal vector is not contained in input features, we estimate a normal vector for each point using open3d~\cite{zhou2018open3d}.
For primitive fitting, input geometric features for each point $i$ contain a coordinate vector $\vec{p}_i$ and a ground-truth normal vector $\vec{n}_i$.
For mobility-based part segmentation task, input features for each point $i$ are composed of a coordinate vector $\vec{p}_i$ and a flow vector $\vec{f}_i$.
The flow vector is estimated according to two input shapes.

\vpara{Baselines.}
For domain-agnostic baselines for general domain generalization problems, like MixStyle~\cite{zhou2021domain}, Meta-learning~\cite{li2018learning}, Gradient Surgery~\cite{Mansilla_2021_ICCV}, we implement them for segmentation tasks carefully with reference to their released code.
For mobility-based part segmentation task, Deep Part Induction~\cite{yi2018deep} is a learning based segmentation network.
Although the test dataset used in our model is the same as the one used in their work, we download the code, re-implement it using PyTorch, and further train it using our training dataset where each pair is generated on-the-fly from inferred meta-data for part mobility information.
Note that the performance reported in the original paper (77.3\% MIoU on the test set) is achieved using several iterations between flow estimation and part segmentation.
The performance of the model using a single iteration is 63.1\% as reported in their work.
It is also different from the one we report, probably due to the different training dataset.
The performance of other baselines such as JLinkage clustering (JLC)~\cite{yuan2016space} and (Spectral Clustering) SC~\cite{tzionas2016reconstructing} are taken from~\cite{yi2018deep} due to the same test datset.

For primitive fitting, SPFN~\cite{li2019supervised} and HPNet~\cite{Yan_2021_ICCV} are two task-specific methods to solve the problem. 
We download the code of SPFN released by the author, carefully re-implement it using PyTorch and test the model on the same train-validation-test split as the one used for our model.
For HPNet, we download the official implementation and test the model's performance on our train-validation-test split. 

For the semantic-based part segmentation, Learning to Group~\cite{luo2020learning} is a two-stage learning-based segmentation network with a representation learning stage and a reinforcement learning based strategy for part segmentation; SGPN~\cite{wang2018sgpn} and GSPN~\cite{yi2019gspn} are also two task-specific segmentation strategies. WCSeg~\cite{kaick2014shape} is a traditional sgementation method.
The performance of those methods are directly taken from~\cite{luo2020learning} due to the same training and test dataset.


\vpara{Software configurations.} 
We use Python 3.8.8 and PyTorch 1.9.1 to write the main code framework.
Other main packages used include torch\_cluster 1.5.9, torch\_scatter 2.0.7, horovod 0.23.0 for PyTorch, etc.

\vpara{Hardware configuration.} 
All training experiments, including supervision search stage and regular training stage, are conducted on 8 NVIDIA Geforce RTX 3090 GPUs in parallel. Experiments for inference stage is conducted on one single NVIDIA Geforce RTX 3090 GPU.

\section{Additional Experimental Results} \label{sec_additional_res}

\subsection{Contrastive Learning based Part Segmentation}
Our main experimental results have proved the effectiveness of HPNet~\cite{Yan_2021_ICCV} on the primitive fitting task.
It is a carefully designed two-stage framework with a representation learning network that learns feature representations, parameters and other geometric features such as normal vectors and further hybrid such learned features for the following Mean-Shift clustering module.
Such design achieves impressive performance on the primitive fitting task, with MIoU performance 79.5\%.
It is not clear whether such network architecture is suitable for other two tasks.
To apply them on the mobility-based part segmentation and semantic-based part segmentation, we abbreviate its first representation learning stage to only learn per-point representations optimized by a contrastive style loss and further fit the optimized representation to the clustering stage.

We conduct experiments by applying such contrastive learning based part segmentation strategy to the mobility-based part segmentation task and the semantic-based part segmentation task using PointNet++~\cite{qi2017pointnetpp} as the backbone.
Results are summarized in Table~\ref{tb_motion_inst_con_loss_optim}.
It can be inferred that although such contrastive based segmentation networks can work well on the primitive fitting task, it cannot get satisfactory results on the mobility-based part segmentation task or semantic-based part segmentation task using PointNet++ as the backbone.
Possible reasons may include 1) Such network architecture is not a universal one for all segmentation tasks, considering that it is originally designed for the primitive fitting task. 2) This network architecture is 
not universally suitable for both PointNet++ and DGCNN.
For comparison, our strategy is more universal compared with the contrastive learning based segmentation strategy.

\begin{table}[htbp]
    \centering
    \caption{\footnotesize
    The performance of the contrastive learning based part segmentation networks (``Contrastive'') on the mobility-based part segmentation task (``Mobility'') and the semantic-based part segmentation task (``Semantic''). ``PN++'' refers to ``PointNet++''. Reported values are performance of the trained networks on out-of-domain test datsets.
    }  
    \vspace{-9pt}
    \begin{tabular}{@{\;}c@{\;}|c|c@{\;}}
    \midrule
        \hline
        \specialrule{0em}{1pt}{0pt} 
        ~ & Mobility & Semantic \\ 
        \cline{1-3} 
        \specialrule{0em}{1pt}{0pt}
        
        Contrastive & 44.0 & 25.9 
        \\ \cline{1-3} 
        
        \gpsss$_\text{PN++}$ & 66.5 & 33.9
        \\ \cline{1-3} 
        \specialrule{0em}{1pt}{0pt}
    \end{tabular} 
    \vspace{-10pt}
    \label{tb_motion_inst_con_loss_optim}
\end{table}

\subsection{Part-aware and Geometry-discriminative Supervision Feature Space}
In our method, we design the supervision space to be aware of part-level information by sampling features from points having the same part labels with the target point.
Moreover, the feature space is made aware of geometric features by using ground-truth geometric features to construct the input feature set.
In this section, we conduct ablation study on the mobility-based part segmentation task to prove the necessity of making the calculated features aware of both part labels and geometric features.

\vpara{Ablating part labels.}
We conduct an experiment to ablate part labels in the calculation of supervision features by using features of points sampled from the neighbourhood of each point without considering whether they belong to the same part. Resulting models are denoted as ``$-$Part Labels'' in Table~\ref{tb_res_motion_part_geometric_abl}.

\vpara{Ablating geometric features.}
We conduct an experiment to ablate geometric features in the supervision feature calculation process by only using part labels from the sampled points as input features in the supervision feature space. Resulting models are denoted as ``$-$Geometric Features'' in Table~\ref{tb_res_motion_part_geometric_abl}.

The results of such ablations are summarized in Table~\ref{tb_res_motion_part_geometric_abl}.
Such results demonstrate the necessity of including geometric features and part labels in the intermediate supervision feature calculation process simultaneously.
Furthermore, being aware of geometric features are more important than including part labels in the supervision feature calculation process if we compare the performance of the resulting models for their respective ablation versions.
Moreover, if we compare the performance of the model ``$-$Geometric Features'' with the model optimized using no intermediate supervisions, it can be discovered that the ablated version, where only part labels are involved in the calculation process,
would result very limited performance improvement.
Possible reasons may include 1) Operators designed for geometric features are not that suitable to encode part labels; 2) Adding intermediate supervisions by only letting the model aware of part labels encoded in another form is not enough to help the model learn more part related cues.
Letting the calculated supervision features aware of part-level geometric information can help the network learn useful cues defining parts for the part segmentation task, thus benefiting the networks' performance better than only using a part of them.

\begin{table}[tbp]
    \centering
    \caption{\footnotesize Ablation study w.r.t. reward function design and supervision space design.  For abbreviations used, ``Arch.'' refers to ``Architecture''; ``PN++'' denotes ``PointNet++''; In/Out-of-dist. refers to In/Out-of-distribution performance. 
    } 
    \vspace{-8pt}
    \begin{tabular}{@{\;}c@{\;}|c|c|c@{\;}}
    \midrule
        \hline
        \specialrule{0em}{1pt}{0pt} 
        Ablation & Arch. & In-dist. & Out-of-dist. \\ 
        \cline{1-4} 
        \specialrule{0em}{1pt}{0pt}
        
        / & \multirow{3}{*}{DGCNN} & 83.1 & \textbf{73.8}
        \\ \cline{1-1} \cline{3-4}
        \specialrule{0em}{1pt}{0pt}
        
        $-$Part Labels & ~ & 82.4 & 71.3
        \\ \cline{1-1} \cline{3-4}
        \specialrule{0em}{1pt}{0pt}
        
        $-$Geometric Features & ~ & 83.9 & 68.9
        \\  \cline{1-4} 
        \specialrule{0em}{1pt}{0pt}
        
    \end{tabular} 
    \vspace{-6pt}
    \label{tb_res_motion_part_geometric_abl}
\end{table}

\subsection{Supervision Feature Generation Strategy}
In this work, we propose to generating supervision features from an operation tree operating on input part-aware geometric features.
No doubt that there are many other methods that can be used to generate such features from ground-truth input features for supervision.
We explore one possibility by letting a network consume such input part-aware geometric features for intermediate ground-truth supervision features. 
The network is optimized together with the main network by gradient descent.
Resulting models are denoted as ``NN'' and the results are summarized in 
Table~\ref{tb_res_motion_network_gen_feat}.
As can be inferred from the table, using a network to generate ground-truth features for prediction is not a wise choice, with the performance improvement that can be added on the original simple segmentation networks very limited.
It indicate that, using such simple gradient descent strategy, a network cannot ``learn'' the correct way to generate high-quality ground-truth features for intermediate supervisions.
This can also prove an effective strategy to use a computing graph-like operation tree for ground-truth features.

\begin{table}[tbp]
    \centering
    \caption{\footnotesize Ablation study w.r.t. reward function design and supervision space design.  For abbreviations used, ``Arch.'' refers to ``Architecture''; ``PN++'' denotes ``PointNet++''; ``No Loss'' indicates simple segmentation networks trained without intermediate supervisions; In/Out-of-dist. refers to In/Out-of-distribution performance.
    } 
    \vspace{-8pt}
    \begin{tabular}{@{\;}c@{\;}|c|c|c@{\;}}
    \midrule
        \hline
        \specialrule{0em}{1pt}{0pt} 
        Ablation & Arch. & In-dist. & Out-of-dist. \\ 
        \cline{1-4} 
        \specialrule{0em}{1pt}{0pt}
        
        / & \multirow{3}{*}{PN++} & 87.2 & \textbf{66.5}
        \\ \cline{1-1} \cline{3-4}
        \specialrule{0em}{1pt}{0pt}
        
        / (No Loss) & ~ & 86.4 & 61.0
        \\ \cline{1-1} \cline{3-4}
        \specialrule{0em}{1pt}{0pt}
        
        NN & ~ & 86.5 & 62.5
        \\  \cline{1-4}  
        \specialrule{0em}{1pt}{0pt}
        
        / & \multirow{3}{*}{DGCNN} & 83.1 & \textbf{73.8}
        \\ \cline{1-1} \cline{3-4}
        \specialrule{0em}{1pt}{0pt}
        
        / (No Loss) & ~ & 88.3 & 68.1
        \\ \cline{1-1} \cline{3-4}
        \specialrule{0em}{1pt}{0pt}
        
        NN  & ~ & 89.5 & 67.0
        \\ \cline{1-4} 
        \specialrule{0em}{1pt}{0pt}
        
    \end{tabular} 
    \vspace{-6pt}
    \label{tb_res_motion_network_gen_feat}
\end{table}

\subsection{Experimental Results Details}
In this section, we present some details of experimental results that are not presented in the main paper.

\begin{figure}[htbp]
    \flushleft
    \includegraphics[width=0.95\linewidth]{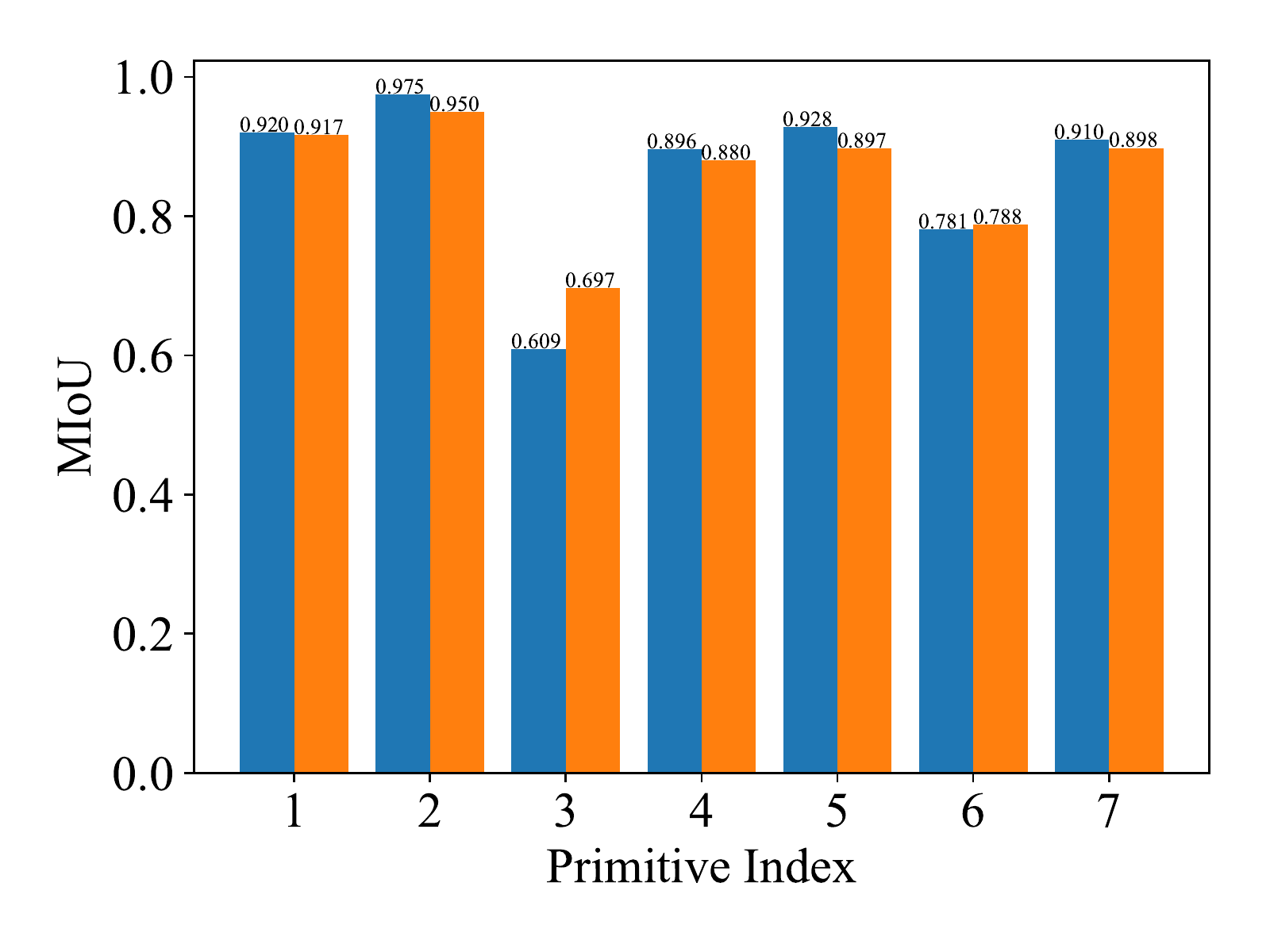}
    \vspace{-7pt}
  \caption{ 
  \footnotesize
  Performance comparison between HPNet and \gpsss$_\text{HPNet}$ on different data clusters.
  }
  \label{fig_perf_comparison_hpnet_and_loss}
\end{figure}

\vpara{Performance comparison between HPNet and \gpsss$_\text{HPNet}$.}
Performance comparison between HPNet and \gpsss$_\text{HPNet}$ on different data clusters is shown in Figure~\ref{fig_perf_comparison_hpnet_and_loss}.

\subsection{Segmentation Results Visualization}
In this section, we visualize and present some segmentation results for the mobility-based part segmentation task and the semantic-based part segmentation task.
Point clouds are normalized into a ball with radius 1.0.
Thus the presented shapes may be different from those used in the training stage.

\begin{figure*}[htbp]
\centering
    
    \subfloat{
        \rotatebox{90}{\scriptsize{~~~~~~~~Frame 2~~~~~~~~~~~~~~~~~~~~~~~~Frame 1}}
        \includegraphics[width=0.9\linewidth]{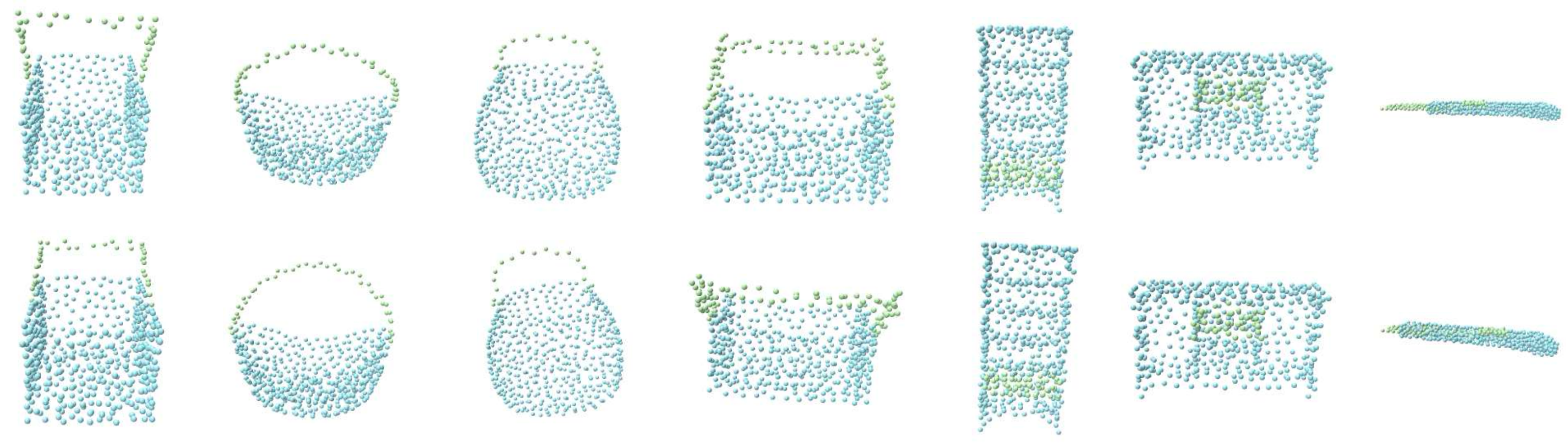}
    }
    
    \noindent\rule{\textwidth}{0.5pt}
    \subfloat{
        \rotatebox{90}{\scriptsize{~~~~~~~~Baseline}}
        \includegraphics[width=0.9\linewidth]{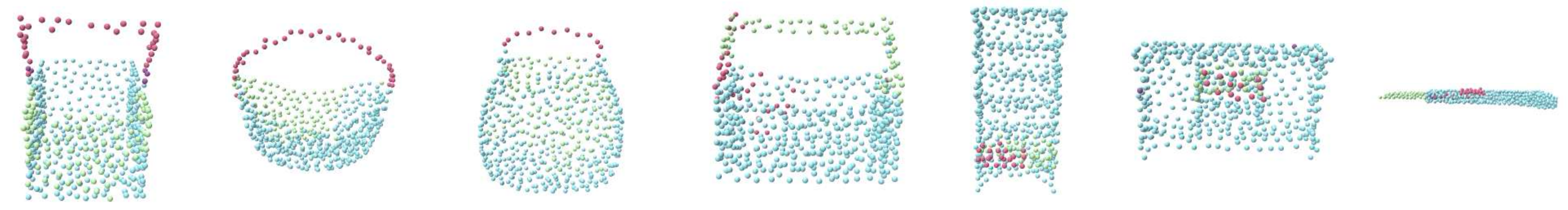}
    }
    
    \subfloat{
        \rotatebox{90}{\scriptsize{~~~~~~~~Ours}}
        \includegraphics[width=0.9\linewidth]{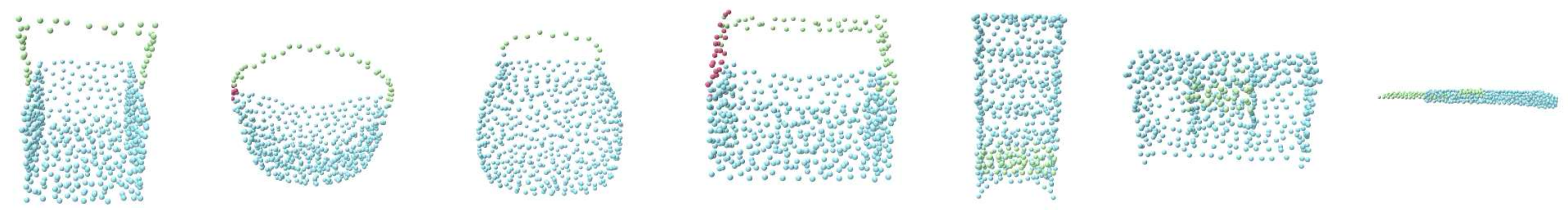}
    }
    
    \subfloat{
        \rotatebox{90}{\scriptsize{~~~~~~~~Frame 2~~~~~~~~~~~~~~~~~~~~~~~~Frame 1}}
        \includegraphics[width=0.9\linewidth]{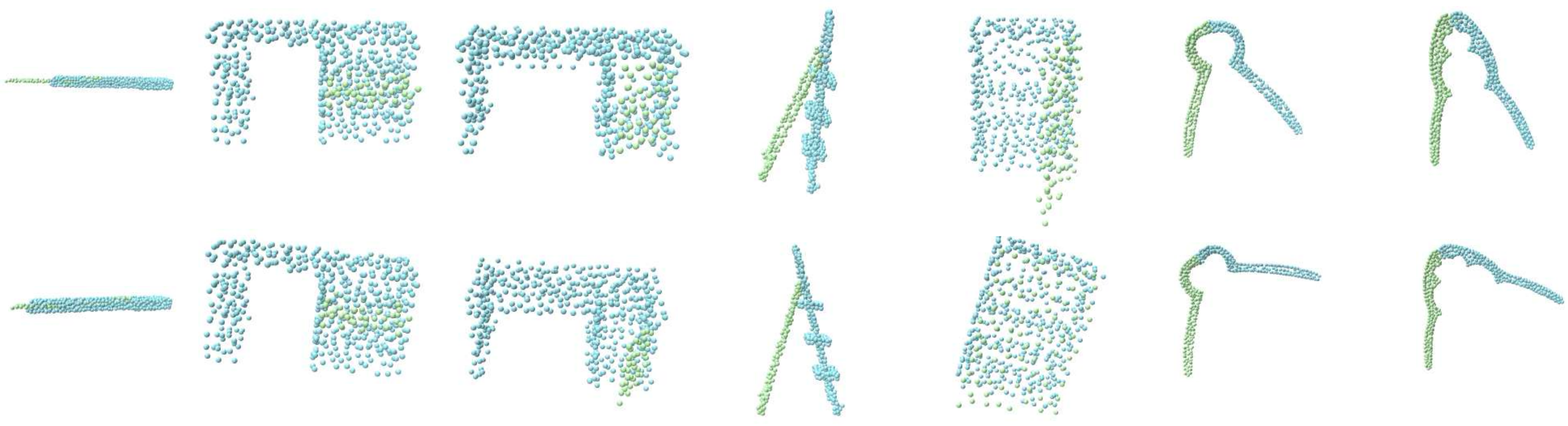}
    }
    
    \noindent\rule{\textwidth}{0.5pt}
    \subfloat{
        \rotatebox{90}{\scriptsize{~~~~~~~~Baseline}}
        \includegraphics[width=0.9\linewidth]{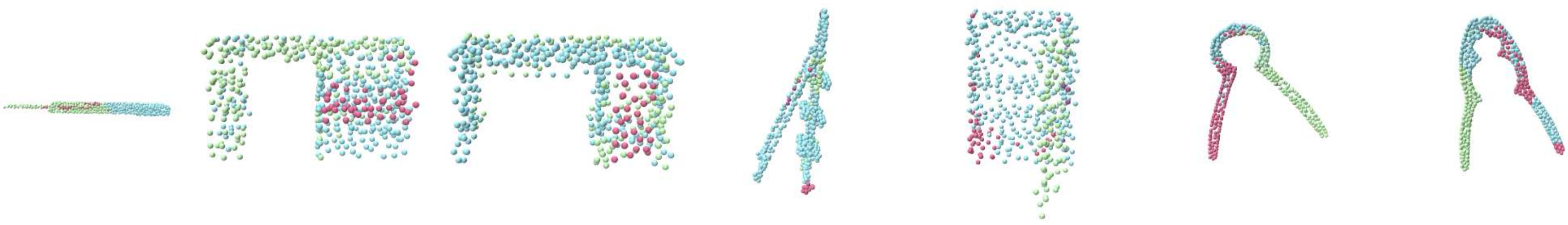}
    }
    
    \subfloat{
        \rotatebox{90}{\scriptsize{~~~~~~~~Ours}}
        \includegraphics[width=0.9\linewidth]{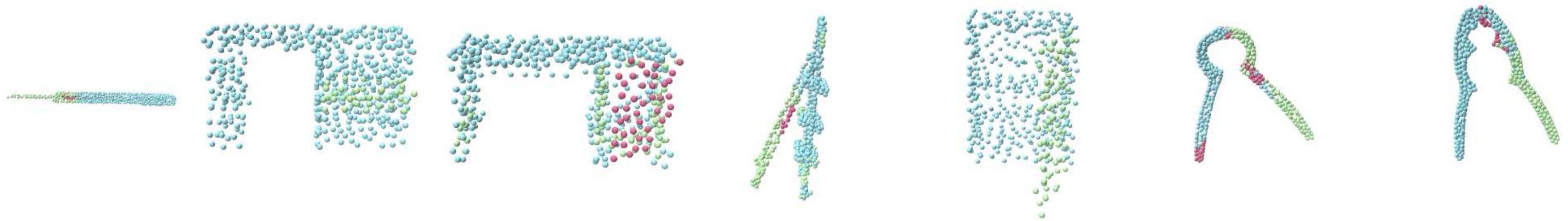}
    }
    
\vspace{-8pt}
  \caption{Mobility-based segmentation results visualization. ``Baseline'' refers to the segmentation network using DGCNN as its backbone without adding any intermedaite supervision. ``Ours'' denotes the model ``\gpsss$_\text{DGCNN}$''.}
  \label{fig_exp_motion_seg_vis_1}
 \vspace{-10pt}
\end{figure*}

\begin{figure*}[htbp]
\centering
    
    \subfloat{
        \rotatebox{90}{\scriptsize{~~~~~~~~Frame 2~~~~~~~~~~~~~~~~~~~~~~~~Frame 1}}
        \includegraphics[width=0.9\linewidth]{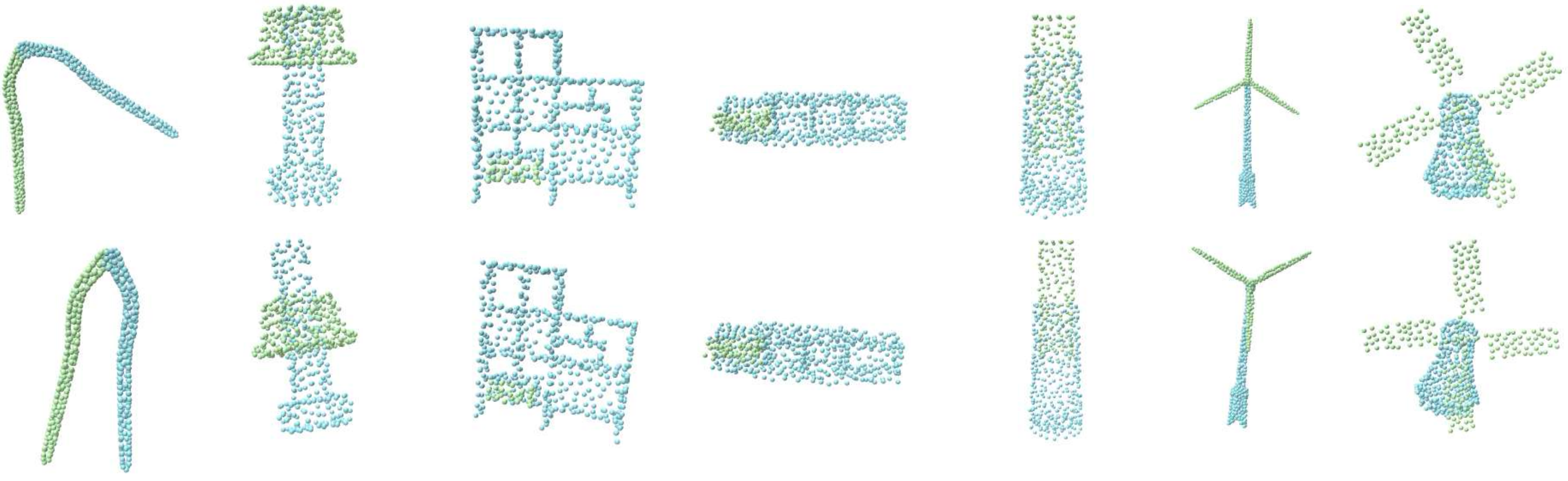}
    }
    
    \noindent\rule{\textwidth}{0.5pt}
    \subfloat{
        \rotatebox{90}{\scriptsize{~~~~~~~~Baseline}}
        \includegraphics[width=0.9\linewidth]{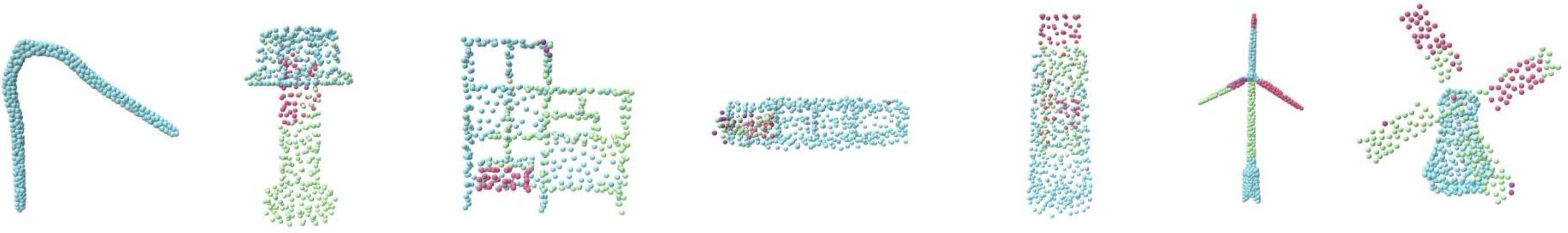}
    }
    
    \subfloat{
        \rotatebox{90}{\scriptsize{~~~~~~~~Ours}}
        \includegraphics[width=0.9\linewidth]{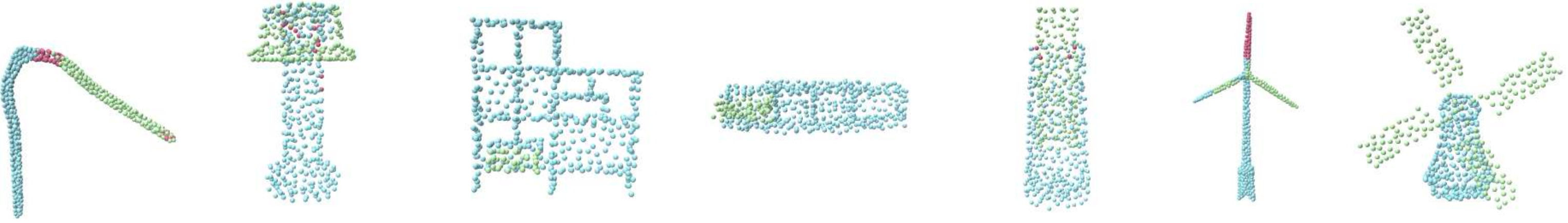}
    }
    
\vspace{-8pt}
  \caption{Mobility-based segmentation results visualization. ``Baseline'' refers to the segmentation network using DGCNN as its backbone without adding any intermedaite supervision. ``Ours'' denotes the model ``\gpsss$_\text{DGCNN}$''.}
  \label{fig_exp_motion_seg_vis_3}
 \vspace{-10pt}
\end{figure*}

\vpara{Mobility-based part segmentation.}
Figure~\ref{fig_exp_motion_seg_vis_1} and~\ref{fig_exp_motion_seg_vis_3} (presented in next few pages)
show the selected segmentation visualization for the mobility-based part segmentation task, where ``Baseline'' denotes ``DGCNN'' model using no intermediate supervision while ``Ours'' denotes ``\gpsss$_\text{DGCNN}$'' model using intermediate loss searched by \gps.



\begin{figure*}[htbp]
\centering
    
    \subfloat{
        \rotatebox{90}{\scriptsize{~~~~~~~~~~~~~~~~Vase G.T}}
        \includegraphics[width=0.9\linewidth]{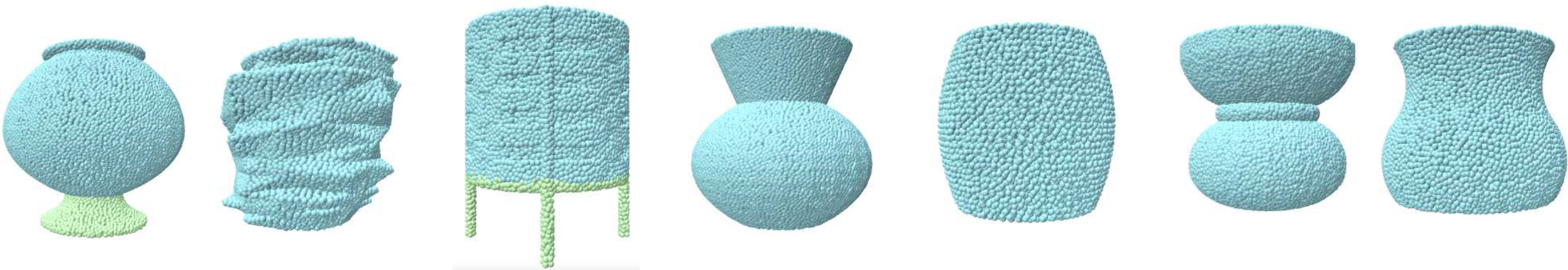}
    }
    
    \subfloat{
        \rotatebox{90}{\scriptsize{~~~~~~~~Vase Pred.}}
        \includegraphics[width=0.9\linewidth]{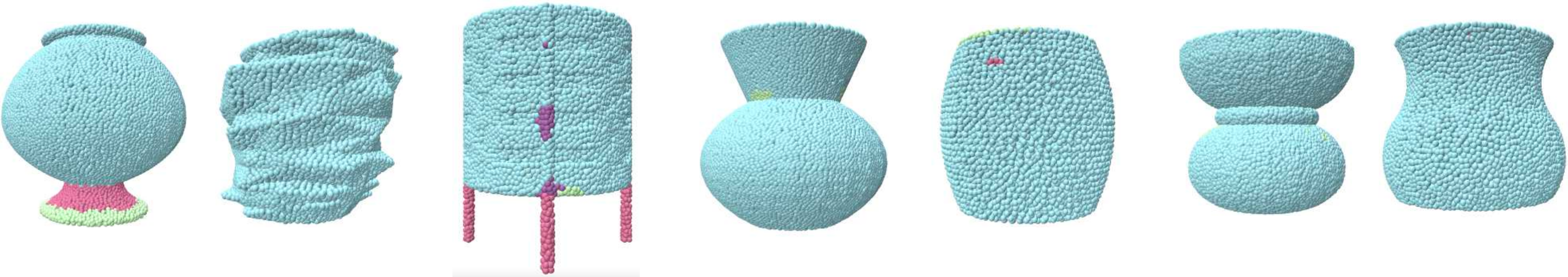}
    }
    
    
    
\vspace{-8pt}
  \caption{Semantic-based segmentation results visualization. ``G.T'' refers to the ground-truth segmentation results. ``Pred.'' denotes the model ``\gpsss''.}
  \label{fig_exp_inst_seg_2}
 \vspace{-10pt}
\end{figure*}
\begin{figure*}[htbp]
\centering
    
    \subfloat{
        \rotatebox{90}{\scriptsize{~~~~~~~~Bag G.T}}
        \includegraphics[width=0.9\linewidth]{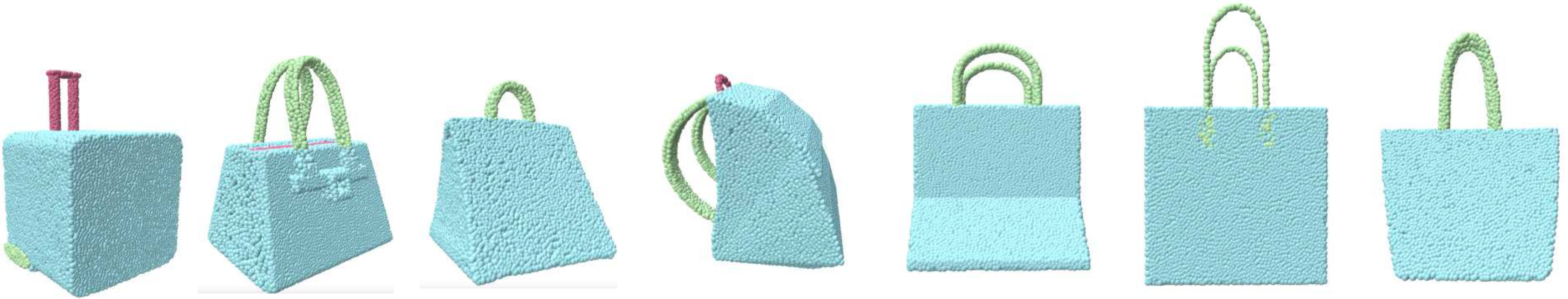}
    }
    
    \subfloat{
        \rotatebox{90}{\scriptsize{~~~~~~~~Bag Pred.}}
        \includegraphics[width=0.9\linewidth]{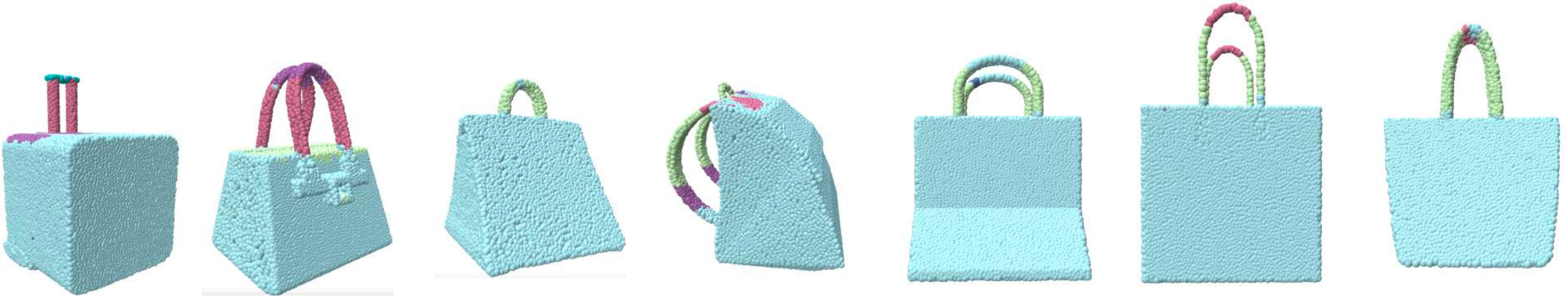}
    }
    
    \subfloat{
        \rotatebox{90}{\scriptsize{~~~~~~~~~~~~Bowl G.T}}
        \includegraphics[width=0.9\linewidth]{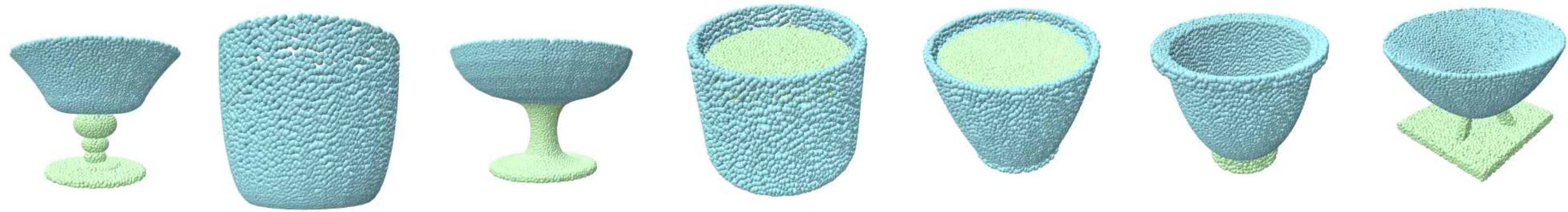}
    }
    
    \subfloat{
        \rotatebox{90}{\scriptsize{~~~~~~~~Bowl Pred.}}
        \includegraphics[width=0.9\linewidth]{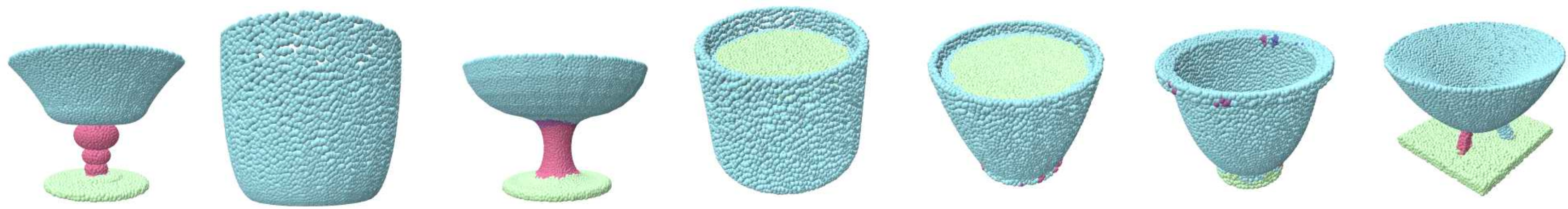}
    }
    
    \subfloat{
        \rotatebox{90}{\scriptsize{~~~~~~~~~~Display G.T}}
        \includegraphics[width=0.9\linewidth]{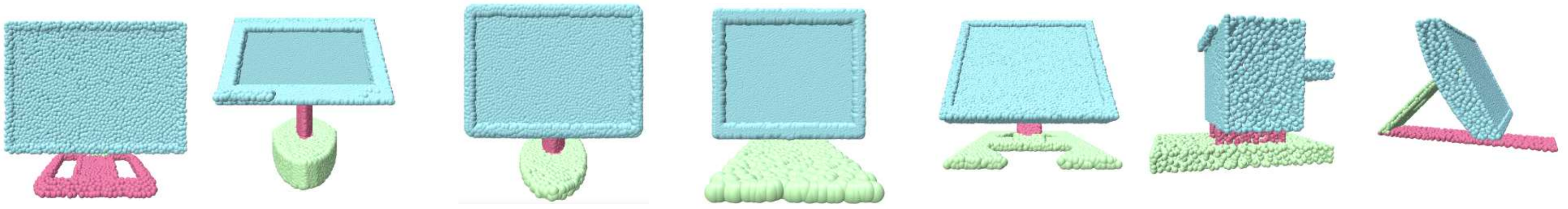}
    }
    
    \subfloat{
        \rotatebox{90}{\scriptsize{~~~~~~~~Display Pred.}}
        \includegraphics[width=0.9\linewidth]{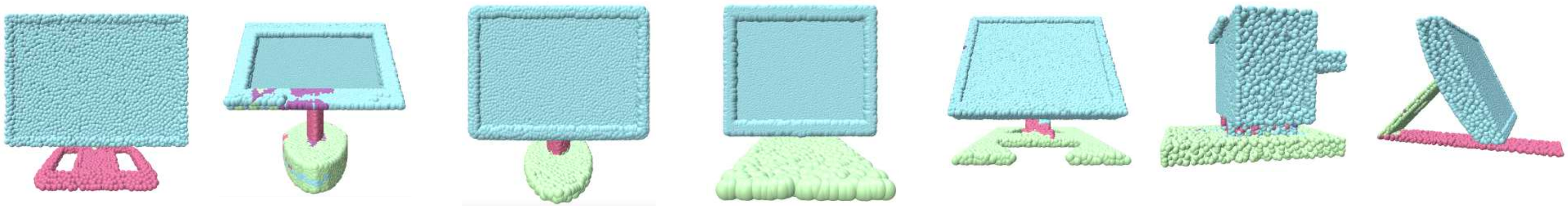}
    }
    
    \subfloat{
        \rotatebox{90}{\scriptsize{~~~~~~~~~~~~~~~~Hat G.T}}
        \includegraphics[width=0.9\linewidth]{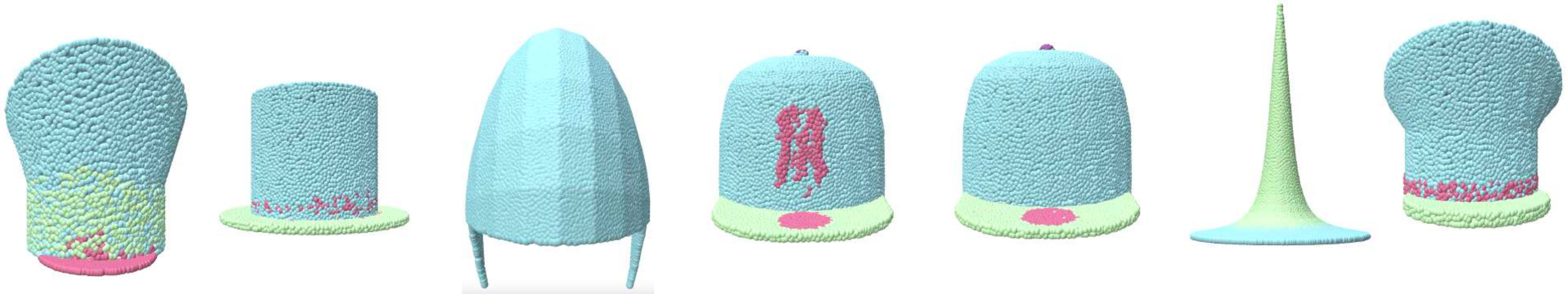}
    }
    
    \subfloat{
        \rotatebox{90}{\scriptsize{~~~~~~~~Hat Pred.}}
        \includegraphics[width=0.9\linewidth]{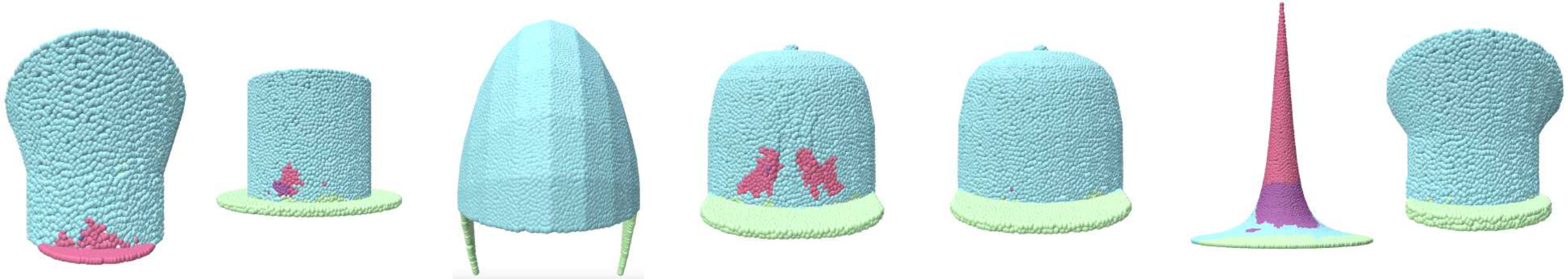}
    }

\vspace{-8pt}
  \caption{Semantic-based segmentation results visualization. ``G.T'' refers to the ground-truth segmentation results. ``Pred.'' denotes the model ``\gpsss''.}
  \label{fig_exp_inst_seg}
 \vspace{-10pt}
\end{figure*}

\vpara{Semantic-based part segmentation.}
Figure~\ref{fig_exp_inst_seg_2} and~\ref{fig_exp_inst_seg} (presented in next few pages) show the segmentation results of \gps on shapes from several test categories for the semantic-based part segmentation task.

\section{Further Discussion}
In this work, we propose to automatically find useful intermediate supervisions to help with improve the generalization ability of 3D part segmentation networks.
Although experiments prove the value of adding intermediate supervisions for improving networks' cross-domain performance, the network structure used in current work is still limited to a single stage end-to-end learning-based network which does not vary across different segmentation networks.
However, it is valuable to explore how to include the architecture of segmentation networks into the design space and automatically select suitable network architectures for different part segmentation tasks.
Making the network architecture flexible can help enlarge the design space, thus including more highly expressive networks into the search space.

Another line lies in more explorations on other tasks, not only limited to 3D part segmentation tasks. It is expected that adding intermediate supervisions a general approach to prevent network from learning shortcut features for tasks from a much broader range. However, it is still not proved and worth further exploration.
Moreover, migrating the research goal from only improving the network's cross-domain generalization ability to enhancing both of its in-distribution and out-of-distribution performance is also a meaningful direction.

\clearpage

\end{document}